\documentclass[runningheads]{IEEEtran}

\usepackage{graphicx}
\usepackage{booktabs}

\usepackage[accsupp]{axessibility}

\usepackage[pagebackref,breaklinks,colorlinks]{hyperref}

\usepackage{xcolor}

\usepackage{orcidlink}

\usepackage{algorithm}
\usepackage{algpseudocode}

\usepackage{xcolor}
\usepackage{ulem}

\definecolor{colorone}{RGB}{255, 0, 0} 
\definecolor{colorcat}{RGB}{30, 110, 220} 
\definecolor{colorrm}{RGB}{127, 127, 127}

\definecolor{colorboying}{RGB}{0, 150, 0}   

\begin{document}

\title{ICM: Intra-class Mixing for Domain Adaptation in Adverse Weather}

\author{Boying Li, Chang Liu, Britta Ayano Wilde, György Kovács, Tosin Adewumi, Björn Backe, and Hamam Mokayed}

\maketitle

\begin{abstract}
Unsupervised domain adaptation (UDA) for semantic segmentation remains challenging under adverse weather conditions because severe appearance changes enlarge the domain gap and degrade the reliability of pseudo labels in the target domain. To address this problem, we propose an Intra-Class Mixing Consistency (ICM) framework that enforces prediction consistency between an intra-class mixed image and its original counterpart. Unlike previous mixing-based consistency methods that combine regions across different images or domains and may introduce unrealistic semantic inconsistencies, ICM performs mixing within the same image and semantic class, preserving realistic semantic layout for consistency regularization. With ICM, we establish a new state-of-the-art performance for clear-to-adverse-weather unsupervised domain adaptation (UDA) in semantic segmentation. On the Cityscapes $\rightarrow$ ACDC benchmark, our method achieves 75.7\% mIoU, outperforming the previous state of the art by +1.9 pp, demonstrating its effectiveness in mitigating class confusion under challenging environmental conditions. The code is provided in the supplementary material.
\end{abstract}

\section{Introduction}
Most large-scale Autonomous Driving (AD) datasets, such as NuScenes~\cite{caesar2020nuscenes}, KITTI\cite{geiger2013vision}, and Waymo~\cite{sun2020scalability}, are predominantly collected under clear weather conditions. Moreover, commonly used off-the-shelf pretrained backbones, such as DINOv3\cite{simeoni2025dinov3}, are also trained primarily on datasets that inherently contain far fewer adverse-weather samples. As a result, many models often struggle to generalize to adverse-weather conditions. Despite their low frequency, these adverse-weather conditions pose significant safety risks. Thus reliable semantic understanding under such conditions is critical for real-world deployment.
A promising step towards this reliable understanding is provided by Unsupervised Domain Adaptation (UDA). UDA uses labeled data from a source domain (e.g., clear weather) and unlabeled data from a target domain (e.g., adverse weather) to improve performance on the target domain without requiring manual annotations. 

However, in the context of AD, the substantial domain gap in semantic segmentation presents serious challenges. Differences in camera setups can introduce viewpoint variations and color tone discrepancies. Furthermore, adverse weather conditions dramatically alter scene appearance: illumination differs significantly between day and night; fog reduces texture contrast and leads to class conflation; raindrops and snowflakes introduce noise and degrade image quality; and snow-covered objects exhibit appearances that deviate considerably from their clear-weather counterparts. Several works attempt to explicitly remove weather induced degradations~\cite{jeong2025robust,chen2020jstasr,li2019heavy,liu2018desnownet,qian2018attentive,qu2019enhanced,quan2019deep,yang2019single,yasarla2019uncertainty}. Other studies mitigate class confusion through cross-domain sample mixing~\cite{Tranheden2021} or incorporate masked image modeling to enhance spatial context reasoning~\cite{hoyer2023mic}. Although UDA methods have made remarkable progress in recent years, a noticeable performance gap compared to fully supervised training still remains.

As highlighted in \cite{hoyer2023mic}, a common issue in UDA is the confusion between visually similar classes in the target domain (e.g., road vs. sidewalk, pedestrian vs. rider), where the absence of ground-truth supervision prevents learning subtle appearance differences. This challenge becomes more pronounced under adverse weather because weather-induced appearance distortions increase the domain gap between source and target images. As observed in MIC\cite{hoyer2023mic}, many existing UDA methods\cite{hoyer2022daformer,hoyer2022hrda}, including MIC, fail to segment snow-covered sidewalk, distinguish sky/vegetation/building in dark nights. Notably, this performance gap remains even when employing state-of-the-art backbones such as DINOv3~\cite{simeoni2025dinov3}, which are pretrained using large-scale self-supervised learning on web-scale data.

To specifically address this challenge, we enforce consistency between predictions on the intra-class mixed image and the clean image, under the assumption that the model’s prediction should remain invariant to within-class appearance permutations. This consistency regularization encourages prediction consistency with intra-class appearance variations.

Implementation-wise, we propose a method that regularizes prediction consistency between intra-class–perturbed images and their clean counterparts. In particular, we introduce an Intra-Class Mixing consistency module (ICM), designed as a plug-in component for UDA-based semantic segmentation. ICM selects confident regions and mixes their pixels into unconfident regions belonging to the same class according to the pseudo-labels generated by an Exponential Moving Average (EMA) teacher\cite{tarvainen2017mean} from the original (i.e., unperturbed) image. 
ICM is a simple consistency module that is easily integrated into existing UDA frameworks. We validate its effectiveness by incorporating it into both MIC~\cite{hoyer2023mic} and DAFormer~\cite{hoyer2022daformer}.

Our method establishes new SOTA (state-of-the-art) performance on clear-to-adverse-weather adaptation benchmarks, as shown in Table~\ref{tab:sota}. Specifically, we set the new SOTA by +1.9 pp compared to the previous SOTA\cite{shen2025w}, and outperforms the proposed baseline by 1.3 - 2 pp (Table~\ref{tab:baseline}) mean IoU on Cityscapes (CS)→ACDC unsupervised domain adaptation task.

\section{Related Work}

\subsection{Unsupervised Domain Adaptation}

Unsupervised domain adaptation (UDA) aims to transfer knowledge from a labeled source domain to an unlabeled target domain by mitigating distribution shifts. For semantic segmentation in autonomous driving, this remains particularly challenging due to variations in weather, illumination, sensor characteristics, and scene layouts, which significantly degrade dense pixel-level predictions~\cite{Tranheden2021}.

Early UDA methods reduce the domain gap by aligning source and target distributions in feature space~\cite{ganin2016domain, hoffman2016fcns, long2018conditional, Tsai2018, sakaridis2025condition}, output space~\cite{saito2018maximum, Tsai2018, vu2019advent, luo2021category}, or through channel-wise feature adaptation~\cite{wu2018dcan} and image-level translation~\cite{yang2020fda, kim2020learning, dundar2018domain}. Although these approaches effectively reduce global distribution discrepancies, alignment alone is often insufficient under large appearance shifts, where pseudo-label quality becomes the main limiting factor. Self-training has therefore become the dominant paradigm for UDA semantic segmentation. Pseudo-labels~\cite{lee2013pseudo} generated on target images can be refined using confidence filtering~\cite{mei2020instance, Zhang_2018_CVPR, Zou2018}, EMA teacher--student frameworks~\cite{tarvainen2017mean}, rare-class sampling~\cite{hoyer2022daformer}, masked consistency learning~\cite{hoyer2023mic}, and uncertainty-aware denoising~\cite{fang2023locating}. 

Adapting from clear-weather domains to adverse-weather domains presents additional challenges, as weather degradations reduce texture information, alter illumination statistics, and increase ambiguity between semantically similar classes. Recent approaches address these issues through masked consistency learning~\cite{hoyer2023mic}, weather-invariant feature alignment~\cite{sakaridis2025condition}, diffusion-based image translation~\cite{shen2025w}, and restoration-guided adaptation~\cite{yang2025semantic}. Our proposed Intra-Class Mixing consistency (ICM) is built upon the standard teacher--student self-training framework by improving pseudo-label robustness through intra-class consistency, and is demonstrated on both DAFormer~\cite{hoyer2022daformer} and MIC~\cite{hoyer2023mic}.

\subsection{Mixing and Consistency Regularization}
Data mixing and consistency regularization have become increasingly important in semantic segmentation and UDA. Based on the region-selection strategy, mixing methods can be categorized into geometric-based approaches\cite{yun2019cutmix}, which employ predefined spatial patterns (e.g., rectangular cutouts), and semantic-aware approaches\cite{olsson2021classmix, Tranheden2021}, where mixing regions are determined according to semantic class boundaries. From a domain-level perspective, mixing strategies can further be divided into naive mixing\cite{olsson2021classmix}, which combines samples within the unlabeled dataset to generate augmented images, and cross-domain mixing\cite{Tranheden2021}, which integrates samples from different domains. At the image level, mixing can be classified into cross-image mixing\cite{yun2019cutmix, olsson2021classmix, Tranheden2021}, where regions from different images are combined, and intra-image mixing\cite{kang2017patchshuffle}, where regions within the same image are rearranged or recomposed.

Early mixing strategies such as CutMix~\cite{yun2019cutmix} perform rectangular region replacement between two images, where a randomly sampled rectangular patch from one image is pasted into another. While effective as a regularizer, CutMix is agnostic to semantic boundaries and may disrupt object structure, as mixed regions do not necessarily align with class semantics. To better preserve semantic consistency, ClassMix~\cite{olsson2021classmix} proposes copying class-specific regions based on predicted segmentation masks, ensuring that pasted areas correspond to coherent semantic objects. By preserving class boundaries, ClassMix improves training stability and segmentation performance compared to purely geometric mixing strategies. 

Cross-domain mixing, such as DACS~\cite{Tranheden2021} and UPC~\cite{fang2023locating}, combines labeled source images with pseudo-labeled target images to generate hybrid samples. To mitigate pseudo-label noise, UPC identifies uncertain target patches and replaces them with corresponding source patches annotated with ground-truth labels. However, UPC employs CutMix~\cite{yun2019cutmix} for patch replacement, where rectangular regions are selected without considering semantic boundaries, potentially disrupting scene semantics and introducing contextual inconsistencies. In contrast, our proposed Intra-Class Mixing (ICM) also performs uncertainty-aware patch-wise mixing but restricts the perturbation to semantically consistent regions within the same target image. This class-preserving strategy regularizes structured pseudo-label errors while maintaining the spatial and semantic coherence of the original scene. DACS is built upon semantic-aware mixing, which mixes source and target images using class masks derived from pseudo-labels, enabling joint supervision from ground-truth source labels and target pseudo-labels. This cross-domain mixing encourages the model to learn from spatially diverse and domain-composed inputs, improving robustness under domain shift. While these methods demonstrate that mixing-based perturbations are effective for regularization and domain adaptation, they operate across different images and often across domains. Consequently, they may introduce contextual inconsistencies, such as vehicles appearing in the sky region in the mixed image. Furthermore, DACS still encounters severe class conflation issues when adapting to adverse-weather target domains. This problem extends beyond commonly confused object classes (e.g., road vs.\ sidewalk, rider vs.\ pedestrian, motorcycle vs.\ bicycle) to background ``stuff'' classes such as road, sidewalk, and sky. These classes typically lack distinctive geometric structures and rely heavily on texture cues, which are substantially degraded or altered under adverse weather. For example, snow-covered terrain and sidewalks are difficult to tell apart, and sidewalks wet by rain can be easily confused with roads. 

Beyond cross-image mixing, intra-image perturbations such as PatchShuffle~\cite{kang2017patchshuffle} and jigsaw-style permutation tasks~\cite{noroozi2016unsupervised} have been explored to improve generalization and representation learning. More recent works employ patch-level transformations to enhance robustness or enforce teacher–student consistency~\cite{cai2023patch}. These approaches demonstrate that perturbing spatial configuration within an image can encourage invariance to local arrangement and reduce over-reliance on fragile cues. Nevertheless, existing intra-image permutation methods are typically class-agnostic and do not explicitly account for structured pseudo-label noise in UDA segmentation.

\section{Method}
\subsection{Intra-class mixing}
\begin{figure*}[tb]
  \centering
  \includegraphics[width=0.8\linewidth]{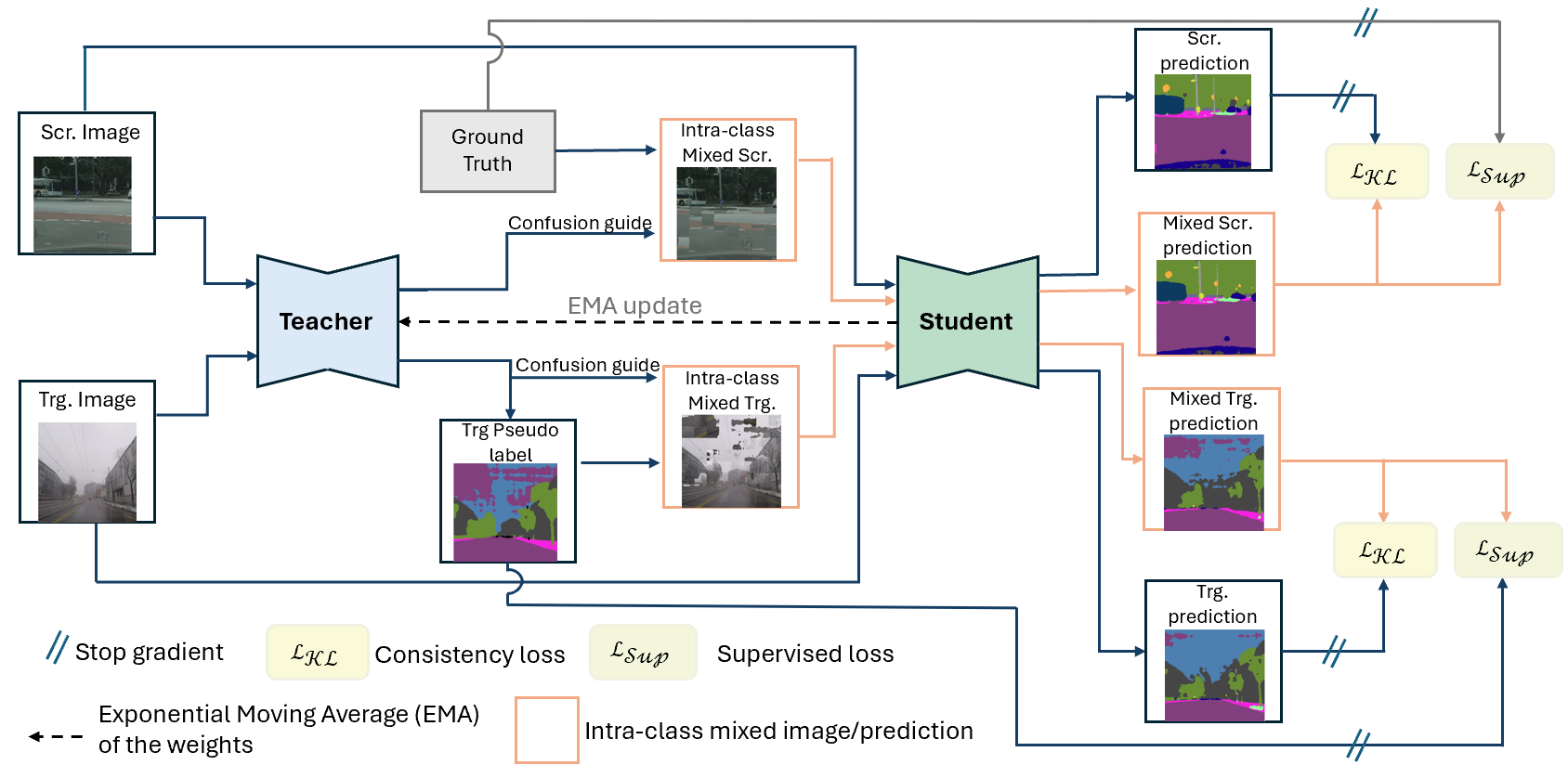}
  \caption{Overview of UDA with the proposed ICM module: Clean images from the source (Src.) and target (Trg.) domains are processed by the teacher model to obtain teacher predictions. For source images, the ground-truth labels provide the semantic class information, while the teacher predictions are used for target images. The teacher's predictions also provide the prediction confidence to guide the mixing direction (Confusion guide \ref{conf-guide}). The consistency between the prediction of the intra-class mixed image and the clean image is enforced by consistency loss and supervision loss.
  }
  \label{fig:overview}
\end{figure*}

\begin{figure*}[tb]
  \centering
  \includegraphics[width=0.8\linewidth]{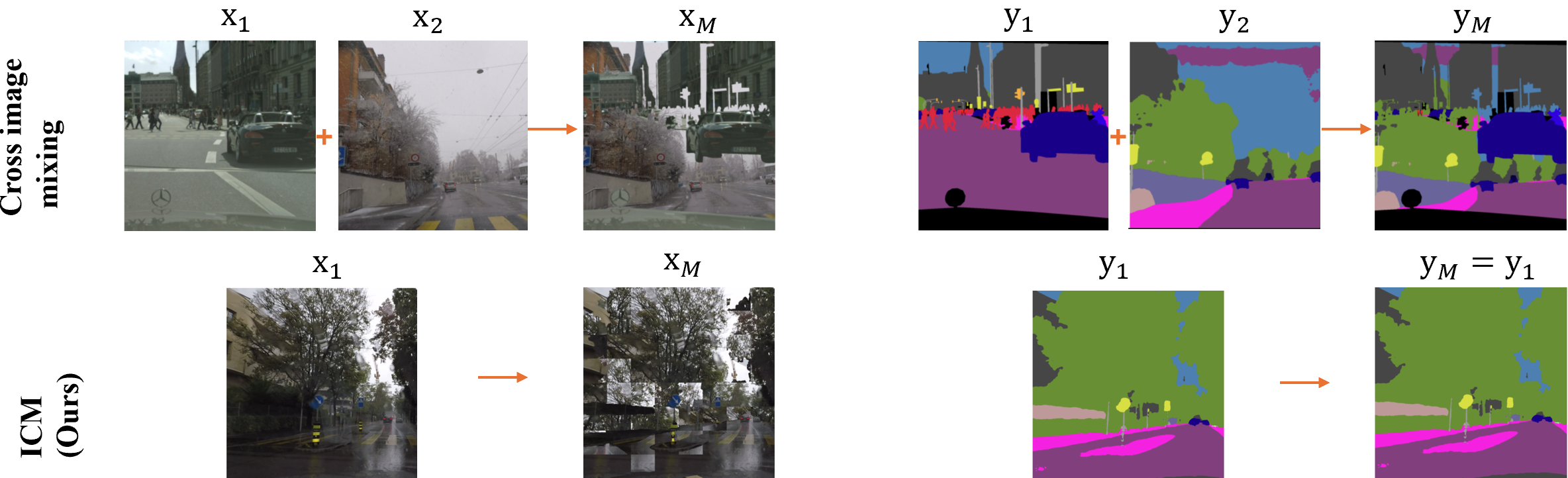}
  \caption{Mixing strategy comparison between cross-image mixing and ICM. x denotes the image, y denotes the corresponding label. $x_M$ and $y_M$ denote the mixed image and the mixed label. ICM performs mixing within the same image and the same class; the label of the intra-class mixed image remains the same as the label of the clean image.
  }
  \label{fig:mix_diag}
\end{figure*}

To address the increased confusion of stuff classes~\cite{Wang2020} in bad weather conditions, we propose Intra-Class Mixing consistency module (ICM) (see Fig.\ref{fig:overview}). ICM mixes regions only within the same image and the same semantic class. Unlike previous cross-image mixing methods that combine different images~\cite{french2019semi,kim2020structured,olsson2021classmix,yun2019cutmix}, ICM perturbs intra-class local patches while preserving the scenes' semantic layout after mixing. This helps the model stay robust to visual changes caused by different weather conditions. The illustration comparing cross-image mixing and ICM is shown in Fig.\ref{fig:mix_diag}. The algorithm of the intra-class mixing process is shown in Algorithm. \ref{alg:intra_mix}

\paragraph{Stuff classes}
Following\cite{Wang2020}, 8 classes out of 19 Cityscapes classes are classified as stuff classes: road, sidewalk, building, vegetation, terrain, wall, fence, and sky.
In this work, we perform the intra-class mixing on the set of stuff classes $\mathcal{C}_{\text{stuff}}$. There are two reasons: (1) in our experiments, we noticed that the conflation under adverse weather is particularly severe for stuff classes~(see Fig.~\ref{fig:qualitative_grid}); (2) compared to object (thing) classes such as people or vehicles, stuff classes generally lack distinctive geometric structures. As a result, disrupting geometric structure is likely less detrimental for stuff classes than for thing classes.

Formally, let $x\in\mathbb{R}^{3\times H\times W}$ denote an input image and $y^*\in\{0,\dots,K-1\}^{H\times W}$ the corresponding pixel labels (ground truth $y$ for source, pseudo-label $\hat{y}$ for target). Following standard UDA practice, a mini-batch is formed by a source image and a target image. For source images, ground truth $y$ is utilized to provide semantic class information.
For target images, pseudo-labels $\hat{y}$ are produced by an EMA teacher to provide semantic information. We define a stuff mask:
\begin{equation}
M_{\mathrm{stuff}}(u) =
\begin{cases}
1 & \text{if } y^*(u) \in \mathcal{C}_{\mathrm{stuff}}, \\
0 & \text{otherwise}.
\end{cases}
\end{equation}
Here, $u$ is the coordinate of an arbitrary pixel, whereas $y^*(u)$ extracts the label value at the corresponding location. For pixels that do not belong to stuff, they will not be permuted.

\paragraph{Patch collection}
 The intra-class mixing is performed on the patch level. Concretely, we partition the image into an $n\times n$ grid of patches. For each class $c\in\mathcal{C}_{\text{stuff}}$, we collect eligible patches whose class occupancy in the patch exceeds a threshold:
\begin{equation}
\mathcal{P}_c
=
\left\{
p \;:\;
\frac{1}{\lvert p \rvert}
\sum_{u \in p}
\mathbf{1}_{\{ y(u) = c \}}
\ge \tau_{\mathrm{patch}}
\right\}.
\end{equation}

\paragraph{Confusion-guided mix.}
To target ``stuff'' regions where the model is uncertain (such as foggy sky, rain-wetted sidewalks, snow-covered terrain), we use the teacher's prediction confidence on the clean image $x$ to guide the mixing direction. Let $p_\theta(\cdot\mid x)$ denote the softmax output from the teacher network on the clean view. For each eligible patch $p\in\mathcal{P}_c$, we compute a confusion score:
\begin{equation}
s(p,c)=\frac{1}{|\Omega_{p,c}|}\sum_{u\in \Omega_{p,c}} \big(1 - (q_1(u)-q_2(u))\big),
\quad \Omega_{p,c}=\{u\in p:\ y(u)=c\},
\end{equation}
where $q_1(u)$ and $q_2(u)$ are the top-2 predicted probabilities at coordinate $u$. We then select high-confusion (``hard'') patches and pair them with low-confusion (``easy'') patches to perform the intra-mixing. Among eligible patches $s(p,c)$, the top 50\% with the highest confusion scores are selected as high-confusion patches, and the other 50\% of eligible patches are considered as low-confusion patches.
The low-confusion patches act as the origin, while the high-confusion patches act as the destination. The pixels from high-confusion pixels will be overwritten by the low-confusion pixels. The overwrite only happens at pixels labeled as class $c$ in both origin and destination locations:
\begin{equation}
\tilde{x}(u) \leftarrow x(\pi(u))\quad \text{for }u\in p_d\text{ such that }y(u)=c\ \wedge\ y(\pi(u))=c,
\end{equation}
where $\pi(\cdot)$ maps destination pixels to origin pixels at the same relative coordinates inside the paired patches. This produces an image $\tilde{x}$ that changes texture/illumination/style within class $c$ while preserving its semantic boundary. The overwritten pixels are recorded in a binary mixing-region mask $M_{\text{mix}}\in\{0,1\}^{H\times W}$, where $M_{\text{mix}}(u)=1$ indicates that pixel $u$ has been replaced by intra-mix. $M_{\text{mix}}$ will be used in the ablative study (Tab. \ref{tab:loss_region}).

\label{conf-guide}

\begin{algorithm}[t]
\caption{Intra-Class Patch Mixing}
\label{alg:intra_mix}
\begin{algorithmic}[1]

\Require Input image $x$, stuff class set $\mathcal{C}$, threshold $\tau$
\Ensure Intra-mixed sample and training loss

\State Divide $x$ into $n \times n$ grid patches

\For{each class $c \in \mathcal{C}$}
    \State Collect patches with class ratio $\geq \tau$
    
    \State Compute confusion score for each patch
    \Statex \hspace{1em} (margin between top-2 predicted probabilities)
    
    \State Select origin patches (low confusion)
    \State Select destination patches (high confusion)
    
    \State Overwrite destination pixels with origin pixels only where both regions belong to class $c$
\EndFor

\State Compute Intra\_mix loss:
\[
\mathcal{L}_{\text{Intra\_mix}}
=
\lambda_{\mathrm{KL}} \mathcal{L}_{\text{KL}}
+
\lambda_{\mathrm{sup}} \mathcal{L}_{\text{sup}}
\]

\label{alg:1}
\end{algorithmic}

\end{algorithm}

\subsection{Intra-Class Mix (ICM) Loss}

We train the segmentation network using an intra-class mixed image $\tilde{x}$ together with its clean counterpart $x$. 
The training objective consists of two components: 
(1) a pixel-wise supervised segmentation loss $\mathcal{L}_{\text{sup}}$, and 
(2) a KL-based consistency regularization $\mathcal{L}_{\text{KL}}$ between clean and mixed predictions.

Rather than having all the regions of an image included in the ICM loss, we can selectively choose the loss region using the semantic masking $M_{\text{stuff}}$ and mixing-region masking $M_{\text{mix}}$. In our framework, we define valid training pixels $\mathcal{R}$ as:
\begin{equation}
    \mathcal{R} = \{u\in \{1,\dots,H\}\times\{1,\dots,W\}| M_{stuff}(u)=1\}.
\end{equation}
More analysis of the loss region can be seen in Section \ref{loss_region_analysis}.

\subsubsection{Supervised Segmentation Loss}

The supervised segmentation loss is computed on the intra-mixed image $\tilde{x}$. 
For each spatial location $u$, we define the effective label $y^*(u)$, representing the ground truth of the source image $y$ and the teacher-generated pseudo-label for the target image $\hat{y}$.

To account for noisy pseudo-labels, we assign pixel weights $W(u)$ such that $W(u)=1$ for valid source pixels, $W(u)=w$ for valid target pseudo-labeled pixels, and $W(u)=0$ for ignored pixels. Following
\cite{hoyer2022daformer,hoyer2022hrda,Tranheden2021,hoyer2023mic},
$w$ is computed as the proportion of target pixels whose maximum
softmax probability exceeds a confidence threshold $\tau$:
\begin{equation}
w =
\frac{1}{HW}
\sum_{i=1}^{H}
\sum_{j=1}^{W}
\mathbf{1}
\!\left[
\max_{c} g_{\phi}(x^{T})_{ijc} > \tau
\right].
\end{equation}
where $g_{\phi}(x^{T})_{ijc}$ denotes the teacher's predicted probability
for class $c$ at pixel $(i,j)$, and $\tau$ is the confidence threshold.

Let $p_\theta(\cdot \mid \cdot)$ denote the corresponding softmax distribution. The supervised loss is defined as a normalized weighted cross-entropy:

\begin{equation}
\mathcal{L}_{\mathrm{sup}}(y^*, \tilde{x}, W)
=
\frac{1}{\sum_{u \in \mathcal{R}} W(u)}
\sum_{u \in \mathcal{R}}
W(u)
\left(
-\log p_\theta(y^*(u)\mid \tilde{x},u)
\right).
\end{equation}

\subsubsection{Intra-Class Mix Consistency}

Beyond supervised learning, ICM enforces prediction consistency between the clean image $x$ and its intra-class swapped version $\tilde{x}$.

Let $\tau_{temp} $ denote a temperature parameter 
and $f_\theta(\cdot) \in \mathbb{R}^{K \times H \times W}$ the segmentation logits, we define the temperature-scaled prediction distribution as
\begin{equation}
p_\theta^{\tau_{temp}}(c \mid x, u)
=
\frac{
\exp\!\left( f_\theta(x)_{c,u} / \tau_{temp} \right)
}{
\sum_{k=1}^{K}
\exp\!\left( f_\theta(x)_{k,u} / \tau_{temp} \right)
}.
\end{equation}

The consistency loss is defined as a pixel-wise KL divergence:

\begin{equation}
\mathcal{L}_{\mathrm{KL}}(x,\tilde{x})
=
\frac{1}{|\mathcal{R}|}
\sum_{u \in \mathcal{R}}
\mathrm{KL}
\!\left(
p_\theta^{\tau_{temp}}(\cdot\mid x,u)
\;\middle\|\;
p_\theta^{\tau_{temp}}(\cdot\mid \tilde{x},u)
\right).
\end{equation}

The clean prediction $p_\theta^{\tau_{temp}}(\cdot \mid x)$ is treated as a fixed target distribution (no gradient), 
while gradients are propagated only through $p_\theta^{\tau_{temp}}(\cdot \mid \tilde{x})$.
This encourages the network to maintain stable semantic predictions under intra-class spatial perturbations.

\subsubsection{Optimization Objective}
The optimization Objective $\mathcal{L}$ of the proposed method consists of two parts, namely the ICM objective $\mathcal{L}_{\mathrm{ICM}}$ and the baseline loss $\mathcal{L}_{\mathrm{base}}$ that we build upon:
\begin{equation}
   \mathcal{L}=  \mathcal{L}_{\mathrm{ICM}}+ \mathcal{L}_{\mathrm{base}}.
\end{equation}
The ICM objective $\mathcal{L}_{\mathrm{ICM}}$ combines supervised segmentation and consistency regularization:

\begin{equation}
\label{eq:total_loss}
\mathcal{L}_{\mathrm{ICM}}
=
\lambda_{\mathrm{sup}}
\mathcal{L}_{\mathrm{sup}}
+
\lambda_{\mathrm{KL}}
\mathcal{L}_{\mathrm{KL}},
\end{equation}
where $\lambda_{\mathrm{sup}}$ and $\lambda_{\mathrm{KL}}$ scale the loss terms.

The baseline objective consists of the supervised source segmentation loss
$\mathcal{L}_{\mathrm{src}}$, the DACS class-mixing loss, and the masking consistency loss,
following DACS~\cite{Tranheden2021}, DAformer\cite{hoyer2022daformer}, HRDA~\cite{hoyer2022hrda},
and MIC~\cite{hoyer2023mic}.

\begin{equation}
\label{eq:base_loss}
\mathcal{L}_{\mathrm{base}}
=
\mathcal{L}_{\mathrm{src}}
+
\mathcal{L}_{\mathrm{dacs}}
+
\mathcal{L}_{\mathrm{mic}}
\end{equation}

\section{Experiments}
\subsection{Implementation Details}

We build ICM upon DAFormer \cite{hoyer2022daformer}, adopting the HRDA multi-resolution self-training framework \cite{hoyer2022hrda} and incorporating the MIC (Masked Image Consistency) module \cite{hoyer2023mic}. ConvNeXt-Base-DINOv3-LVD1689M encoder \cite{simeoni2025dinov3} is used as the backbone. To preserve the pretrained representations, the first two stages of the encoder are frozen during UDA training. Our implementation adopts the standard teacher–student self-training framework of DAFormer, including confidence-weighted pseudo-labels, rare-class sampling \cite{hoyer2022daformer}, and target-domain data augmentation following DACS \cite{Tranheden2021}. The EMA teacher is updated with momentum $\alpha = 0.999$, following \cite{hoyer2022daformer,hoyer2022hrda,hoyer2023mic}. We use the AdamW optimizer \cite{loshchilov2017decoupled} with a learning rate of $6 \times 10^{-5}$ for the encoder and $6 \times 10^{-4}$ for the decoder, linear learning-rate warm-up, and a mini-batch size of $B=2$. For intra-class mixing, each image is divided into a grid $n \times n$ with $n=8$, resulting in 64 patches per image. The KL divergence uses a temperature of 1.0. ICM loss scale $\lambda_{\mathrm{sup}}$ and $\lambda_{\mathrm{KL}}$ are set to 0.5. 
All experiments were conducted on a server equipped with four NVIDIA A100-SXM4-80GB GPUs 
(80\,GB memory each). Training was performed using a single GPU per experiment.

\subsection{Datasets}
In our experiments, we focus on the setting of domain adaptation
and generalization from normal to adverse weather conditions,
as our method particularly tackles the class conflation challenge under adverse weather.
We use Cityscapes\cite{cordts2016cityscapes} as the labeled source-domain set in
our experiments. It contains 2,975 training images, 500 validation images, and 1,525 test images collected during clear weather conditions. In our UDA experiments, source-domain samples are from the Cityscapes training split.

As the target domain with adverse conditions, we adopt the ACDC dataset \cite{sakaridis2021acdc}, which is designed to valuate capabilities on adapting from normal to challenging weather scenarios. ACDC includes 4,006 street-scene images evenly distributed across four adverse conditions: night, fog, rain, and snow. The dataset is divided into 1,600 training images, 406 validation images, and 2,000 test images. In our setup, we use the ACDC training split as the unlabeled target data and evaluate the validation split for ablation studies and hyperparameter tuning.

\subsection{Comparison to the State of the Art in UDA}
MIC\cite{hoyer2023mic} with MiT-B5 backbone\cite{xie2021segformer} has served as a strong baseline in unsupervised domain adaptation (UDA) for semantic segmentation. However, recent advances in large-scale pretrained vision backbones, such as DINOv3\cite{simeoni2025dinov3}, have substantially improved representation learning capabilities. By integrating a DINOv3 backbone into the MIC framework, we establish a significantly stronger baseline (denoted as MIC-DINO). Building upon this enhanced foundation, our proposed ICM further improves adaptation performance by a substantial margin.
\begin{table*}[t]
\centering
\caption{Comparison of state-of-the-art unsupervised domain adaptation methods on Cityscapes→ACDC(Test). The first, second and third groups of rows present DeepLabv2-based UDA, SegFormer-based UDA and diffusion-based UDA, respectively.}
\resizebox{\textwidth}{!}{
\begin{tabular}{lcccccccccccccccccccc}
\hline
Method 
& road & sidew. & build. & wall & fence & pole & light & sign 
& veget. & terrain & sky & person & rider & car & truck & bus 
& train & motorc. & bicycle & mIoU \\
\hline

AdaptSegNet\cite{Tsai2018}
& 69.4 & 34.0 & 52.8 & 13.5 & 18.0 & 4.3 & 14.9 & 9.7 
& 64.0 & 23.1 & 38.2 & 38.6 & 20.1 & 59.3 & 35.6 & 30.6 
& 53.9 & 19.8 & 33.9 & 33.4 \\

BDL\cite{Li2019}
& 56.0 & 32.5 & 68.1 & 20.1 & 17.4 & 15.8 & 30.2 & 28.7 
& 59.9 & 25.3 & 37.7 & 28.7 & 25.5 & 70.2 & 39.6 & 40.5 
& 52.7 & 29.2 & 38.4 & 37.7 \\

CLAN\cite{Luo2019}
& 79.1 & 29.5 & 45.9 & 18.1 & 21.3 & 22.1 & 35.3 & 40.7 
& 67.4 & 29.4 & 32.8 & 42.7 & 18.5 & 73.6 & 42.0 & 31.6 
& 55.7 & 25.4 & 30.7 & 39.0 \\

CRST\cite{Zou2019}
& 51.7 & 24.4 & 67.8 & 13.3 & 9.7 & 30.2 & 38.2 & 34.1 
& 58.0 & 25.2 & 76.8 & 39.9 & 17.1 & 65.4 & 3.7 & 6.6 
& 39.6 & 11.8 & 8.6 & 32.8 \\

FDA\cite{yang2020fda}
& 73.2 & 34.7 & 59.0 & 24.8 & 29.5 & 28.6 & 43.3 & 44.9 
& 70.1 & 28.2 & 54.7 & 47.0 & 28.5 & 74.6 & 44.8 & 52.3 
& 63.3 & 28.3 & 39.5 & 45.7 \\

SIM\cite{Wang2020}
& 53.8 & 6.8 & 75.5 & 11.6 & 22.3 & 11.7 & 23.4 & 25.7 
& 66.1 & 8.3 & 80.6 & 41.8 & 24.8 & 49.7 & 38.6 & 21.0 
& 41.8 & 25.1 & 29.6 & 34.6 \\

MRNet\cite{Zheng2021}
& 72.2 & 8.2 & 36.4 & 13.7 & 18.5 & 20.4 & 38.7 & 45.4 
& 70.2 & 35.7 & 5.0 & 47.8 & 19.1 & 73.6 & 42.1 & 36.0 
& 47.4 & 17.7 & 37.4 & 36.1 \\

DACS\cite{Tranheden2021} 
& 58.5 & 34.7 & 76.4 & 20.9 & 22.6 & 31.7 & 32.7 & 46.8 
& 58.7 & 39.0 & 36.3 & 43.7 & 20.5 & 72.3 & 39.6 & 34.8 
& 51.1 & 24.6 & 38.2 & 41.2 \\

\hline

DAFormer\cite{hoyer2022daformer}
& 58.4 & 51.3 & 84.0 & 42.7 & 35.1 & 50.7 & 30.0 & 57.0 
& 74.8 & 52.8 & 51.3 & 58.2 & 32.6 & 82.7 & 58.3 & 54.9 
& 82.4 & 44.1 & 50.7 & 55.4 \\

SePiCo\cite{xie2023sepico} 
& 61.3 & 48.6 & 84.9 & 39.6 & 40.3 & 54.2 & 48.9 & 60.6 
& 74.8 & 54.3 & 57.2 & 65.2 & 38.3 & 84.8 & 66.2 & 60.4 
& 85.5 & 44.5 & 53.1 & 59.1 \\

HRDA\cite{hoyer2022hrda} 
& 88.3 & 57.9 & 88.1 & 55.2 & 36.7 & 56.3 & 62.9 & 65.3 
& 74.2 & 57.7 & 85.9 & 68.8 & 45.6 & 88.5 & 76.4 & 82.4 
& 87.7 & 52.7 & 60.4 & 68.0 \\

MIC\cite{hoyer2023mic}
& 90.8 & 67.1 & 89.2 & 54.5 & 40.5 & 57.2 & 62.0 & 68.4
& 76.3 & 61.8 & 87.0 & 71.3 & 49.4 & 89.7 & 75.7 & 86.8 
& 89.1 & 56.9 & 63.0 & 70.4 \\

CISS\cite{sakaridis2025condition}
& 92.0 & 69.6 & 89.2 & 57.3 & 40.5 & 55.8 & 67.1 & 67.3 
& 75.3 & 59.7 & 86.4 & 70.0 & 47.5 & 88.9 & 73.1 & 77.5 
& 87.0 & 55.6 & 61.7 & 69.6 \\

HALO\cite{franco2023hyperbolic}
& 94.2 & \underline{79.8} & 88.2 & 60.2 & \underline{51.1} & \textbf{64.1} & \textbf{78.2} & 65.6 
& \textbf{87.9} & 55.7 & \underline{95.5} & 66.3 & 20.7 & 88.9 & \textbf{82.2} & \textbf{89.3}
& 87.9 & 50.4 & 59.0 & 71.9 \\

CoDA\cite{gong2024coda}
& 93.1 & 72.7 & \underline{90.7} & 57.3 & 47.4 & 56.8 & 69.9 & 70.0 
& \underline{87.3} & 59.8 & 95.4 & 71.4 & 47.6 & 90.3 & 77.1 & 83.8 
& 89.1 & 54.7 & \underline{64.1} & 72.6 \\

WA2Net\cite{pan2025exploring} 
& \textbf{94.5} & 77.1 & \textbf{91.6} & \underline{61.9} & 47.6 & 57.5 & 68.5 & \underline{70.4}
& 79.2 & \underline{64.0} & \textbf{95.6} & \underline{72.8} & \underline{50.2} & \underline{90.5} & 75.5 & \underline{88.7}
& \underline{90.9} & \underline{57.5} & 62.6 & \underline{73.8} \\

\hline
W-ControlUDA\cite{shen2025w}
&\textbf{94.5} &77.5 &90.6 &59.1 &44.6 &59.8& 69.8& 61.2& 86.1 & 61.7& 94.1& 70.5& 49.9 &90.1& 75.0& 87.8& 89.3 & 51.4& 59.6& 72.8 \\
\hline

ICM (ours)
&90.2 &\textbf{83.5} & 86.9& \textbf{68.1} & \textbf{54.3} &\underline{63.2} & \underline{73.5} & \textbf{74.1} &78.1 & \textbf{70.4} & 77.4 & \textbf{77.2} & \textbf{60.4} & \textbf{92.3}  &\underline{78.4} & 87.5 &\textbf{93.0} & \textbf{62.0} & \textbf{67.2}& \textbf{75.7} \\
\hline

\end{tabular}
}
\label{tab:sota}
\end{table*}

Tab. \ref{tab:sota} presents the comparison between ICM and existing state-of-the-art UDA methods on the Cityscapes $\rightarrow$ ACDC test benchmark. ICM consistently outperforms all competing methods in terms of mean IoU, achieving a performance gain exceeding 1.9 pp over previous approaches. Across individual categories, ICM achieves the best or second-best IoU in 14 out of 19 classes, including those critical for autonomous driving perception, such as sidewalk, traffic light, traffic sign, person, and various vehicle classes.

\subsection{Analysis and Ablative Study}
We perform in-depth analysis and ablate the components of ICM. The analysis is performed with DAFormer\cite{hoyer2022daformer}, without HRDA \cite{hoyer2022hrda} and MIC \cite{hoyer2023mic},  on Cityscapes(CS)→ACDC. This setting not only enables faster training but also demonstrates that ICM provides consistent and robust improvements on its own, independent of the MIC module~\cite{hoyer2023mic}. Each experiment in the ablative study is trained for 20k iterations and repeated over three independent runs.

\begin{table*}[t]
\centering
\begin{minipage}{0.35\linewidth}
\centering

\centering
\caption{ICM with DAformer~\cite{hoyer2022daformer} for images from different domain. CS$\rightarrow$ ACDC (Val)}
\label{tab:ICM_domain}
\begin{tabular}{lccc}
\toprule
ICM Domain &  mIoU & std\\
\midrule
--             & 62.6& 0.6\\
Source         & 63.0& 0.9\\
Target         & 64.2&0.6\\
Source+Target  & 64.5& 0.7\\
\bottomrule
\end{tabular}

\end{minipage}
\hfill
\begin{minipage}{0.63\linewidth}

\centering
\caption{ICM ablation study with DAFormer~\cite{hoyer2022daformer} on CS$\rightarrow$ACDC(Val).}
\label{tab:icm_ablation}
\begin{tabular}{c c c c c c c }
\toprule
 Exp.& Mixing.  & $\mathcal{L}_{\mathrm{Sup}}$ & $\mathcal{L}_{\mathrm{KL}}$ & Conf-guide & mIoU & std \\
\midrule
1 & -- & -- & --  & -- &62.6&0.6 \\
2 & $\checkmark$ &  $\checkmark$ & $\checkmark$ & $\checkmark$ & 64.5&0.7 \\

3 & Inter-class & $\checkmark$ &  --& -- & 60.5&0.2 \\
4 & $\checkmark$ & $\checkmark$ &  $\checkmark$  & -- & 63.9&1.2 \\

5 & $\checkmark$  & -- &  $\checkmark$ &  $\checkmark$& 63.4&1.3 \\
6 & $\checkmark$  & $\checkmark$ &  -- & $\checkmark$ & 64.3&1.1 \\

\bottomrule
\end{tabular}
\label{tab:ablative}
\end{minipage}
\end{table*}

\subsubsection{Apply ICM to source or target}

Table~\ref{tab:ICM_domain} presents the performance of applying the ICM module to images from different domains: (1) source only, (2) target only, and (3) both source and target. Applying ICM solely to the source domain yields a modest improvement of +0.4 pp. In contrast, applying ICM to the target domain results in a larger gain of +1.6 pp. Incorporating ICM into both the source and target domains achieves the best performance, with an improvement of +1.9 pp.

\subsubsection{Ablation of components}

To gain a deeper understanding of the ICM module, the core components of ICM are ablated in Tab.\ref{tab:ablative}. First, we examine the importance of preserving semantic layout of the scene. Row 3 corresponds to inter-class mixing, where the mixing still happens within the same image, but the mixing is across different classes. In this setting, the labels of inter-class mixed images are updated accordingly, meaning that overwritten destination pixels are assigned the source class ID. 

Since our KL loss assumes that the mixed image should match the clean prediction within the same semantic region, it is not applicable to inter-class mixing where the semantics change. Therefore, only the supervised loss is used in this case. We observe a significant performance drop compared to baseline (-2.1 pp) with inter-class mixing, indicating that preserving semantic identity is crucial for effective training. Compared to random mixing, Confusion-guided mixing increases the mIoU performance by 0.6 pp (Rows 2 and 4).
The supervised loss is a crucial objective for training ICM. Using only the KL loss results in a +0.8 pp improvement over the baseline (Rows 1 and 5), whereas using only the supervised loss achieves a larger gain of +1.7 pp (Rows 1 and 6). With the complete ICM, the mIoU is increased by +1.9 pp (Rows 1 and 2).

\subsubsection{Loss region}
\label{loss_region_analysis}
\begin{table*}[t]
\centering
\caption{ICM loss region study with DAFormer~\cite{hoyer2022daformer} on CS$\rightarrow$ACDC(Val).}
\label{tab:loss_region}
\begin{tabular}{c c c c c c c}
\toprule
 Exp. & Non-Mixed-stuff & Mixed-stuff & Thing & mIoU$_{CS\rightarrow ACDC (Val)}$ & std \\
\midrule
1 & -- & -- & --   & 62.6&0.6 \\
2 & $\checkmark$ & -- &  -- & 64.5&0.7 \\
3 & -- & $\checkmark$ & -- &  64.4 &0.9 \\

4 & $\checkmark$ & $\checkmark$ & --  &  64.5&0.7\\
5 & -- & -- & $\checkmark$  &  62.1&0.4\\

\bottomrule
\end{tabular}
\end{table*}

We further ablate the effect of restricting the intra-mix loss to different regions of the image in Tab.\ref{tab:loss_region}. The image can be defined as three regions: non-mixed-stuff represents pixels whose semantic class belongs to stuff classes and were not changed by the mix; mixed-stuff means the stuff class pixels that were overwritten by the mix; and thing means pixels whose semantic class is not stuff classes.
We observe that the performance does not change much as long as the ICM losses are applied within the stuff classes regions (row 3-5). However, when the losses are applied to thing region only, the performance is even lower than the baseline (row 1 and row 2). This degradation suggests that intra-mix consistency is more suitable for stuff regions, while applying it exclusively to thing regions may introduce harmful perturbations, possibly because object regions are less stable under mixing and more sensitive to pseudo-label noise. 

\subsubsection{Patch size}
In Tab.\ref{tab:patch_size}, we analyze the effect of the grid patch size $n$. Large patches overly disturb the global layout within stuff regions, while very small patches only affect local texture and provide weak appearance variation. The 8×8 grid balances structural coherence and effective intra-class invariance. Performance remains stable across 4–16 grids, indicating ICM is not overly sensitive to this hyperparameter.
\begin{table}[t]
\centering
\caption{ICM patch size study with DAFormer~\cite{hoyer2022daformer} on CS$\rightarrow$ACDC(Val).}
\label{tab:patch_size}
\begin{tabular}{c c c c c c c}
\toprule
&n(Grid size) &-- & 4 & 8 & 16 \\
\midrule
&mIoU$_{CS\rightarrow ACDC (Val)}$  & 62.6& 63.8 & 64.5 & 63.5 \\
&std  & 0.6& 1.4 & 0.7 & 1.1 \\

\bottomrule
\end{tabular}
\end{table}

\subsection{Effectiveness of ICM}
\begin{table*}[t]
\centering
\caption{Comparison of MIC and ICM on Cityscapes→ACDC.}
\resizebox{\textwidth}{!}{
\begin{tabular}{lcccccccccccccccccccc}
\hline
Method 
& road & sidew. & build. & wall & fence & pole & light & sign 
& veget. & terrain & sky & person & rider & car & truck & bus 
& train & motorc. & bicycle & mIoU \\
\hline
\multicolumn{21}{c}{MiT-B5 backbone Cityscapes$\rightarrow$ACDC (Val)} \\
\hline
MIC-Reproduce
& \underline{67.7} & \underline{62.3} & \textbf{80.8} & \textbf{54.0} & \underline{44.3} & \textbf{61.1} & \underline{64.6} & \textbf{62.3}
& \underline{75.8} & \textbf{41.1} & \underline{60.7} & \textbf{67.7} & \textbf{45.6} & \textbf{88.6} & \underline{80.0} & \underline{87.6}
& \underline{86.7} & \textbf{48.4} & \textbf{56.2} & \underline{65.0} \\

ICM(MiT-B5)
& \textbf{87.7} & \textbf{72.9} & \underline{78.8} & \underline{52.2} & \textbf{47.3} & \underline{54.0} & \textbf{68.4} & \underline{57.3}
& \textbf{76.2} & \underline{41.0} & \textbf{78.2} & \underline{60.3} & \underline{43.8} & \underline{88.3} & \textbf{80.6} & \textbf{88.7}
& \textbf{90.5} & \underline{45.1} & \underline{51.0} & \textbf{66.5} \\
\hline
\multicolumn{21}{c}{DINO backbone Cityscapes$\rightarrow$ACDC (Val)} \\
\hline
MIC(DINO)
& \underline{85.0}& \underline{84.6}& \textbf{89.3}& \textbf{66.0}& \textbf{56.1}& \textbf{69.8} & \underline{80.3}& \underline{69.8}&
 \textbf{84.1}& \underline{53.5}& \underline{82.8}& \textbf{75.1}& \underline{54.8}& \underline{91.9}& \underline{85.1}& \underline{92.1}&
 \textbf{93.7}& \textbf{62.3}& \underline{62.6} & \underline{75.7} \\

ICM(DINO)
& \textbf{93.4}& \textbf{85.8} & \underline{88.9}& \underline{65.9} & \underline{55.9}& \underline{68.4}& \textbf{81.2}& \textbf{70.0}&
 \underline{83.7}& \textbf{55.7}& \textbf{87.7}& \underline{74.2} & \textbf{56.9}& \textbf{92.2}& \textbf{89.5}& \textbf{94.7}&
 \underline{93.1}& \underline{62.1}& \textbf{64.6}& \textbf{77.0}\\
\hline
\multicolumn{21}{c}{DINO backbone Cityscapes$\rightarrow$ACDC (Test)} \\
\hline
MIC(DINO)
& \underline{69.3} & \underline{82.2} & \underline{84.1} & \textbf{68.4} & \underline{52.1} & \textbf{63.7} & \textbf{74.3} & \underline{73.0}
& \textbf{86.6} & \underline{66.4} & \underline{60.4} & \underline{76.3} & \textbf{60.8} & \textbf{92.5} & \textbf{78.9} & \textbf{88.9}
& \textbf{93.5} & \textbf{62.7} & \underline{65.6} & \underline{73.7} \\

ICM(DINO)
& \textbf{90.2} & \textbf{83.5} & \textbf{86.9} & \underline{68.1} & \textbf{54.3} & \underline{63.2} & \underline{73.5} & \textbf{74.1}
& \underline{78.1} & \textbf{70.4} & \textbf{77.4} & \textbf{77.2} & \underline{60.4} & \underline{92.3} & \underline{78.4} & \underline{87.5}
& \underline{93.0} & \underline{62.0} & \textbf{67.2} & \textbf{75.7} \\
\hline
\end{tabular}
}
\label{tab:baseline}
\end{table*}

\begin{figure*}[t]
\centering
\setlength{\tabcolsep}{1pt}   
\renewcommand{\arraystretch}{1.0}

\begin{tabular}{ccccc}
\textbf{Image} & \textbf{MIC-DINO} & \textbf{ICM(ours)} & \textbf{Ground Truth} \\

\includegraphics[width=0.2\textwidth]{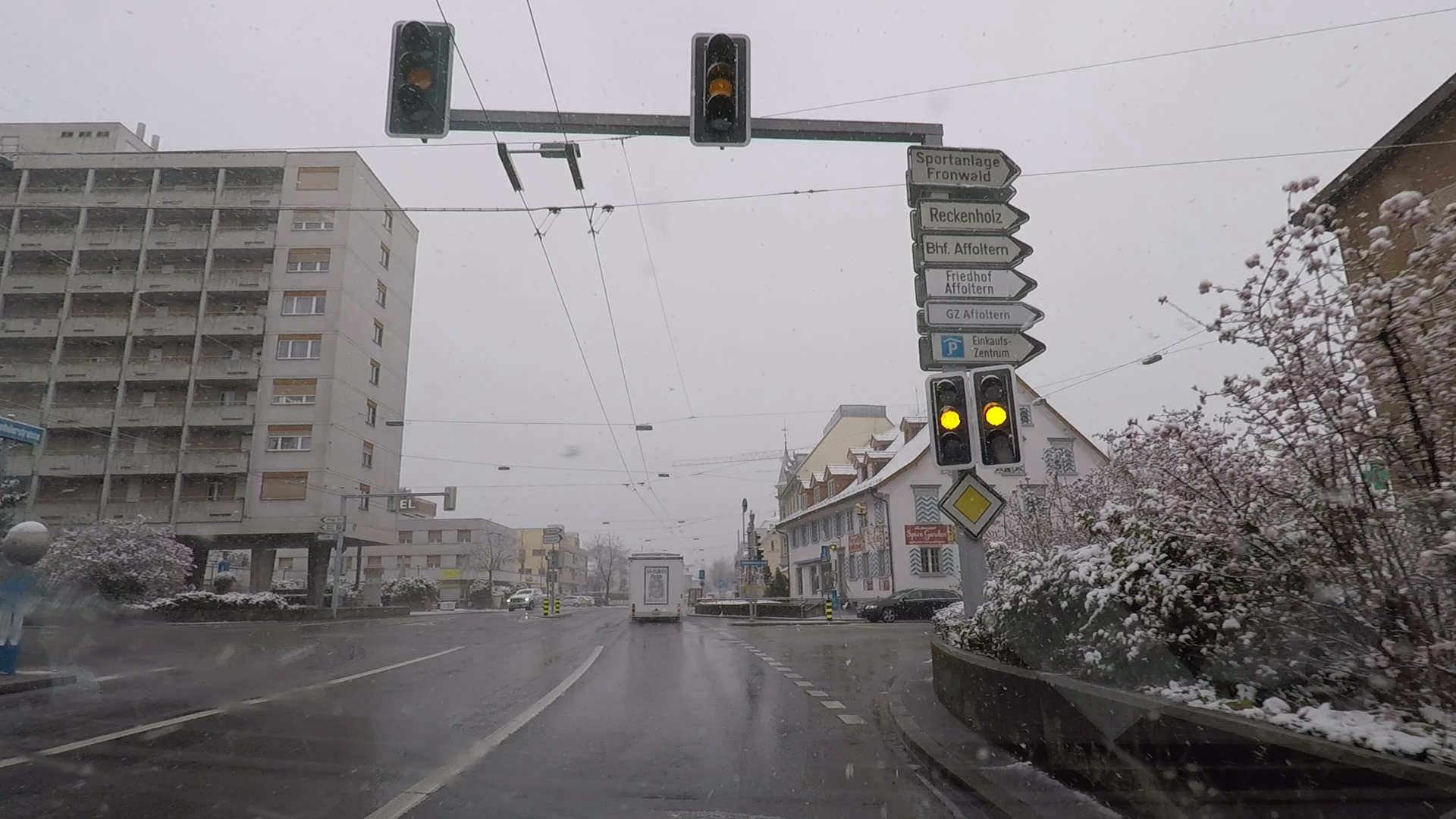}
& \includegraphics[width=0.2\textwidth]{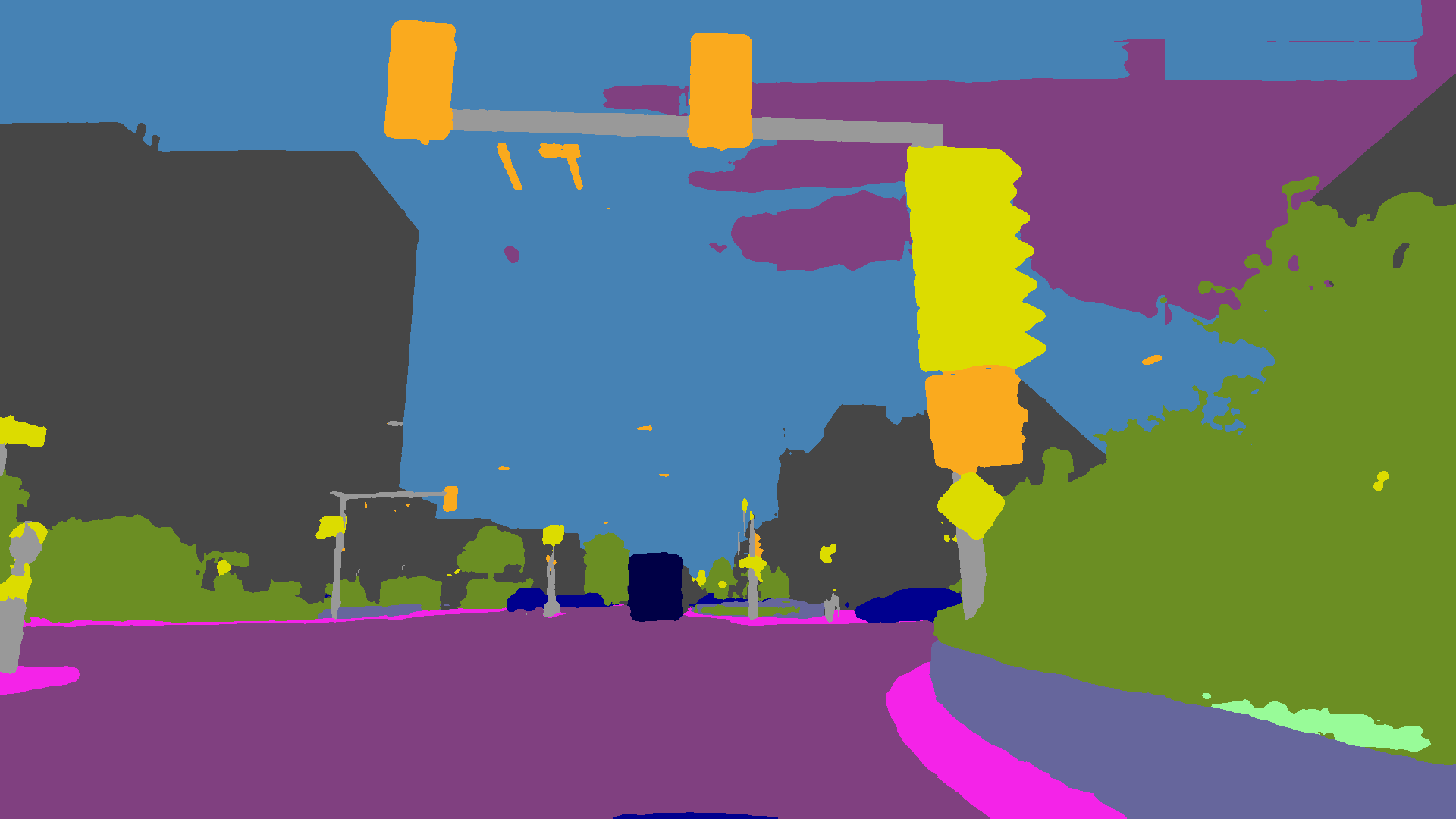}
& \includegraphics[width=0.2\textwidth]{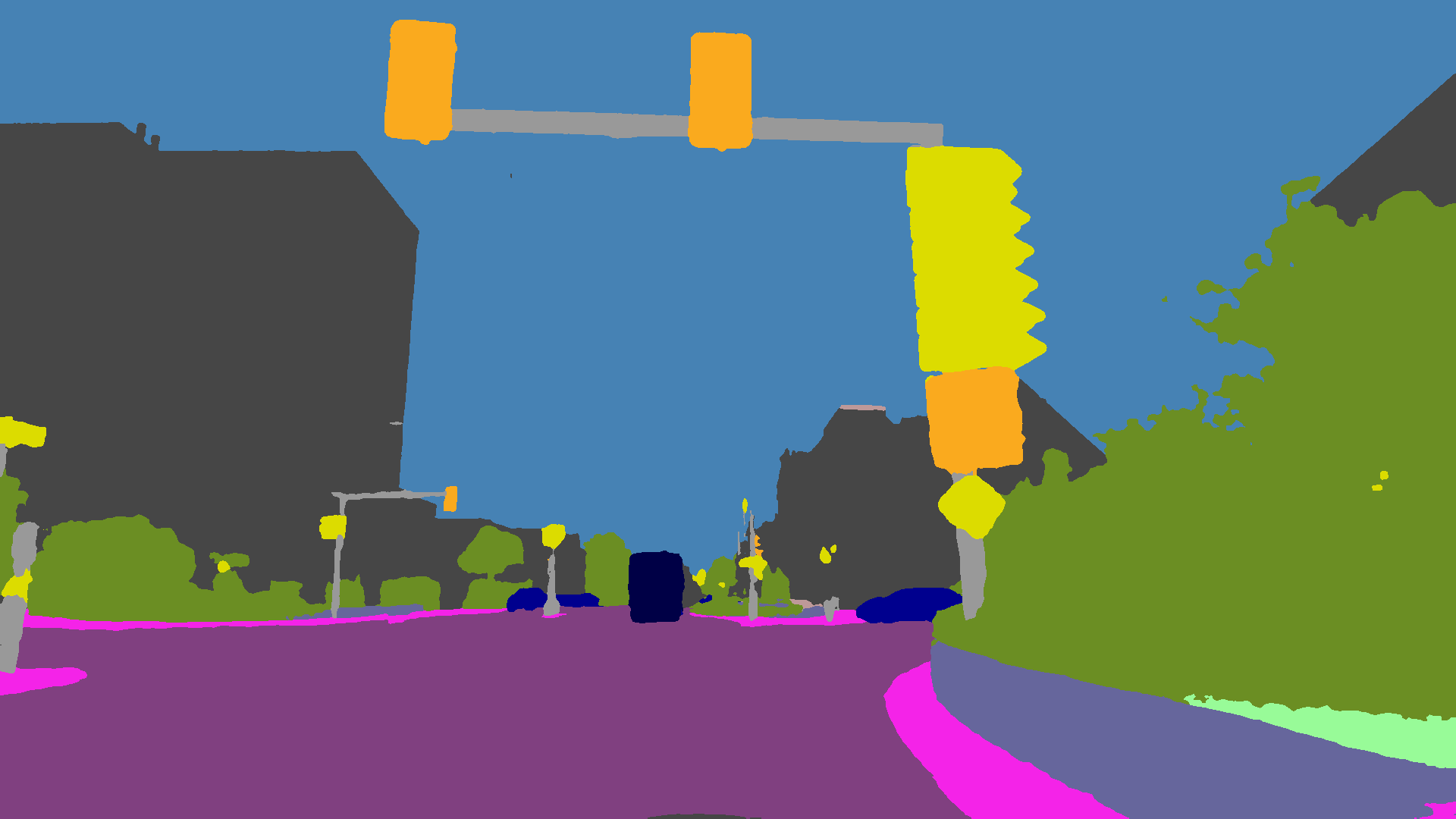}
& \includegraphics[width=0.2\textwidth]{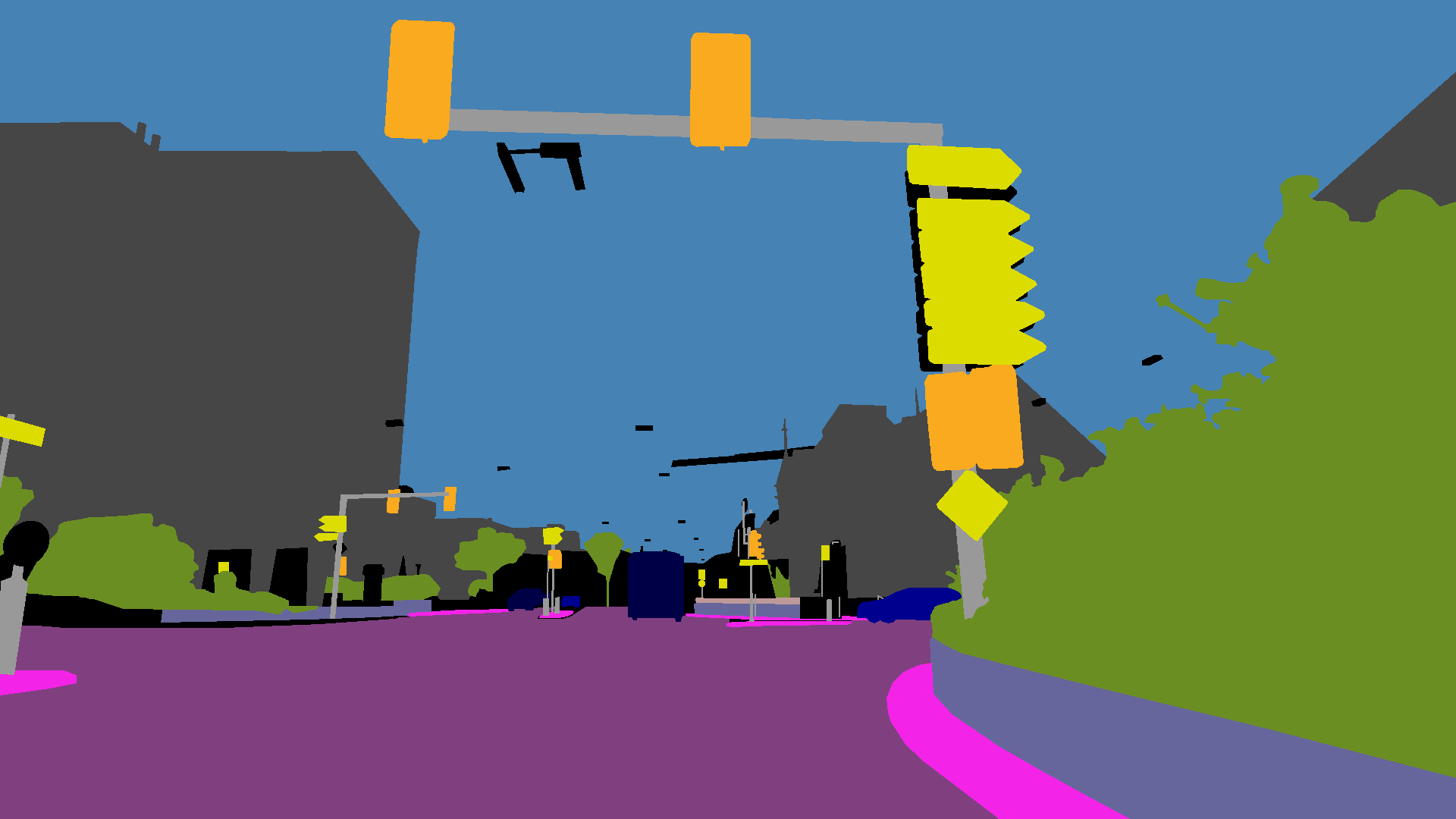} \\

\includegraphics[width=0.2\textwidth]{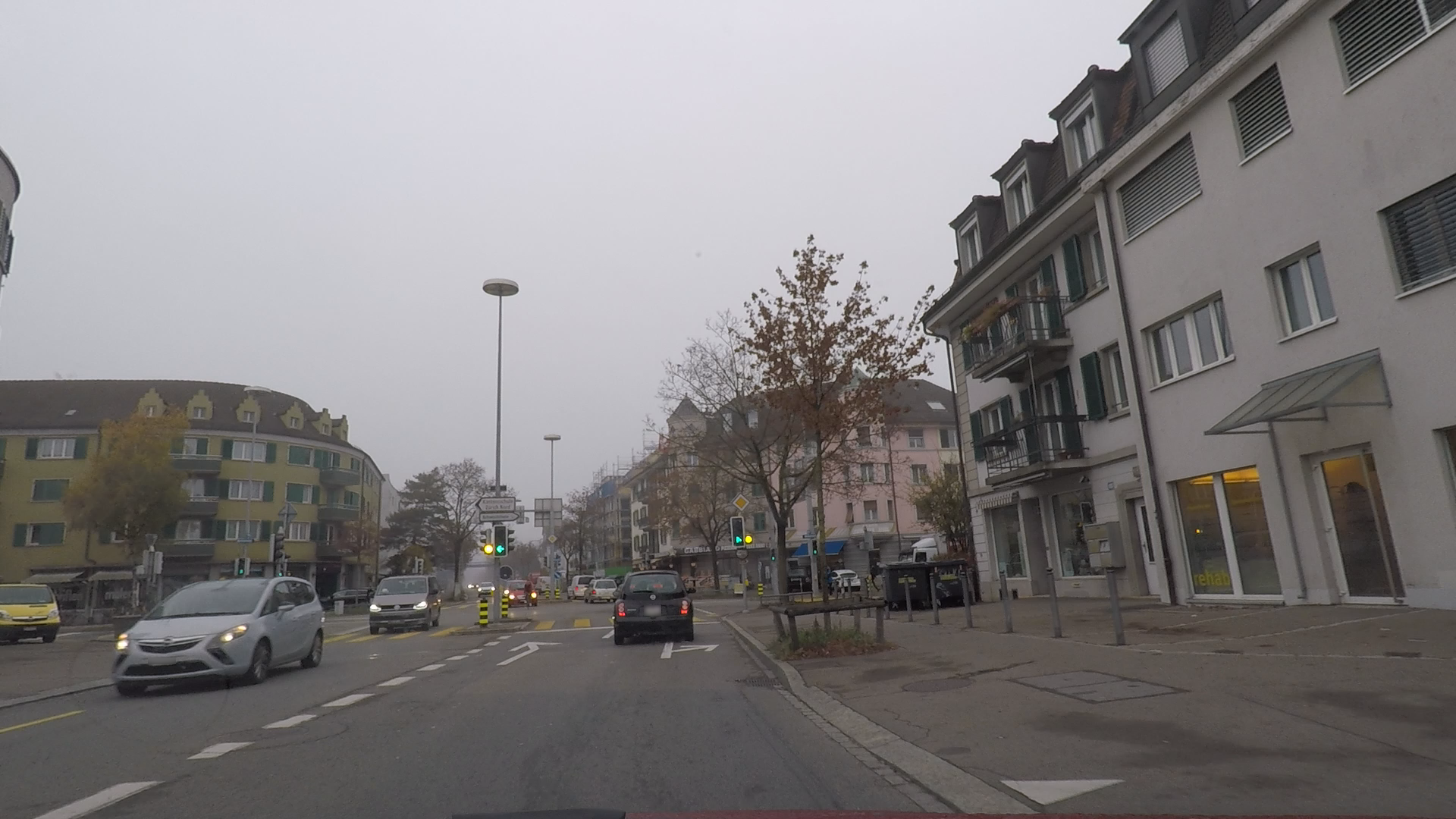}
& \includegraphics[width=0.2\textwidth]{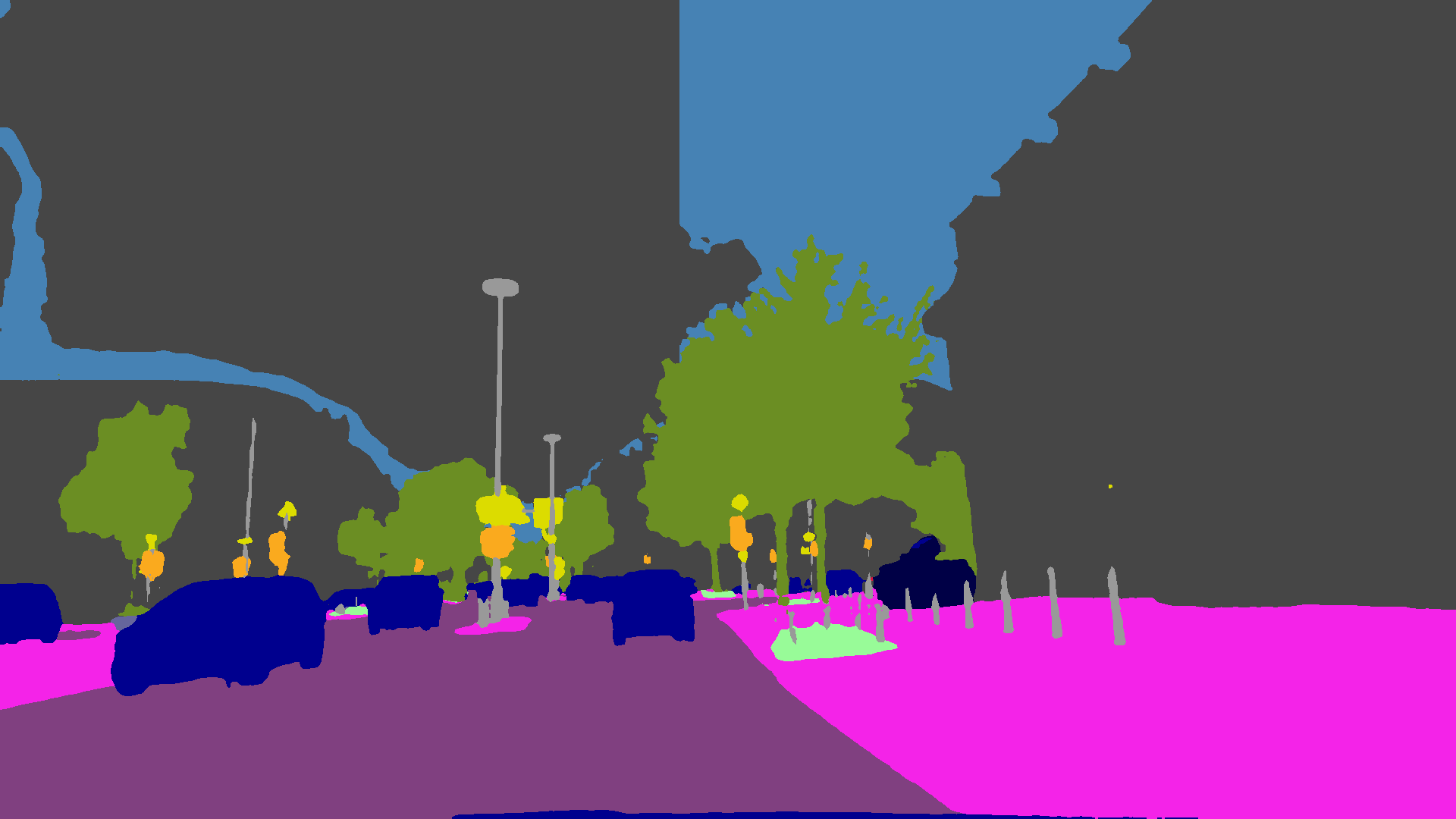}
& \includegraphics[width=0.2\textwidth]{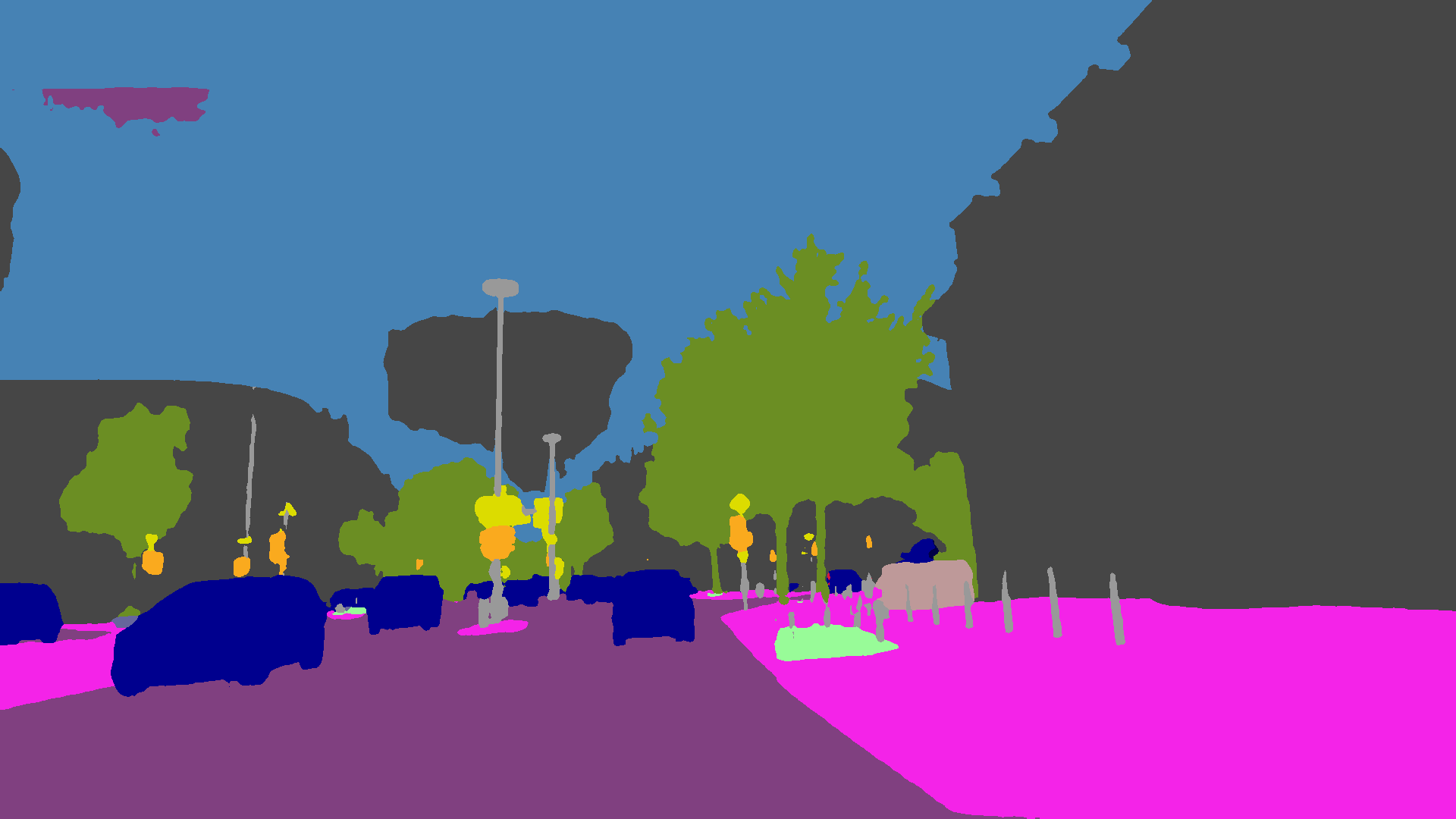}
& \includegraphics[width=0.2\textwidth]{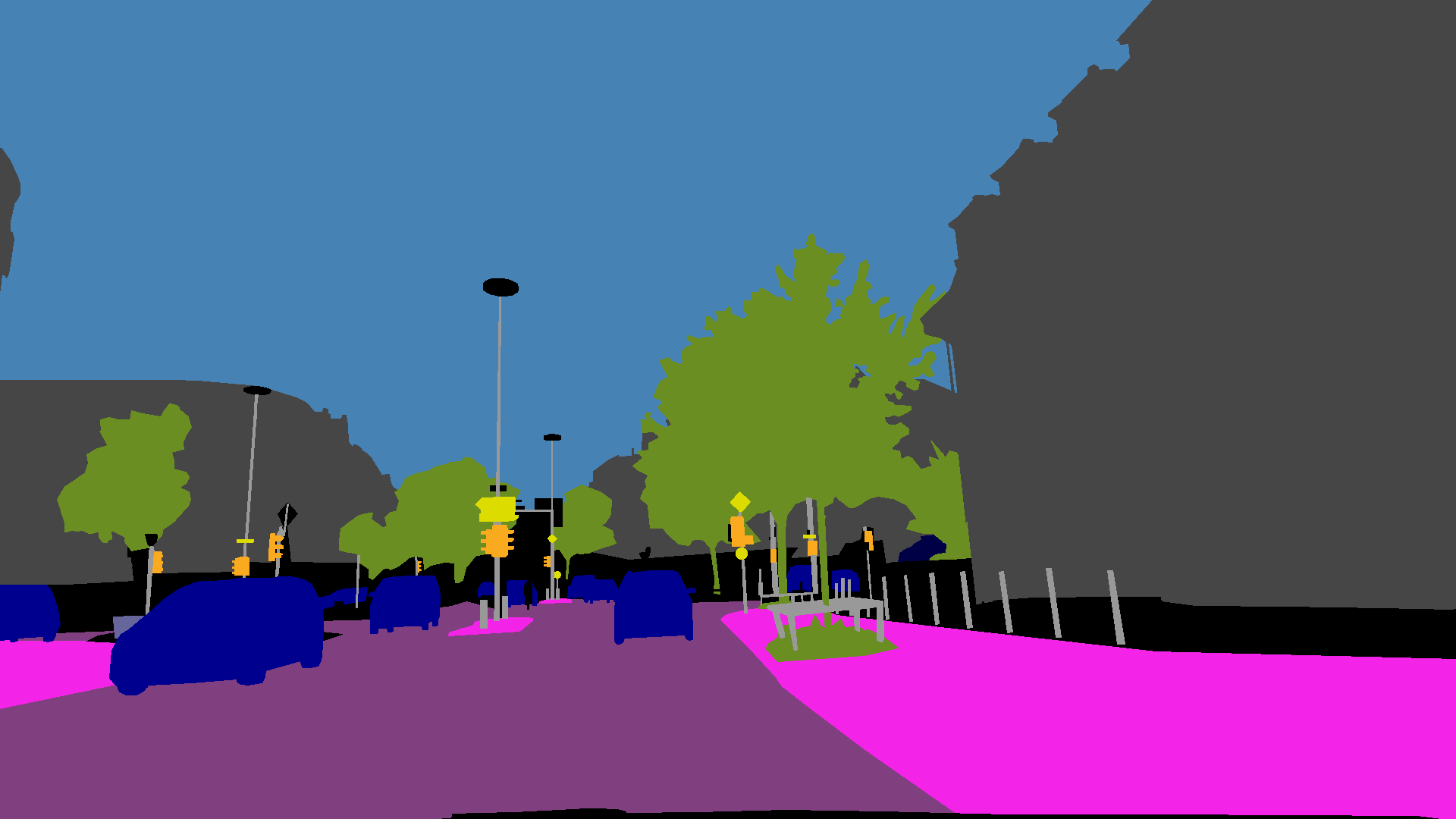} \\

\includegraphics[width=0.2\textwidth]{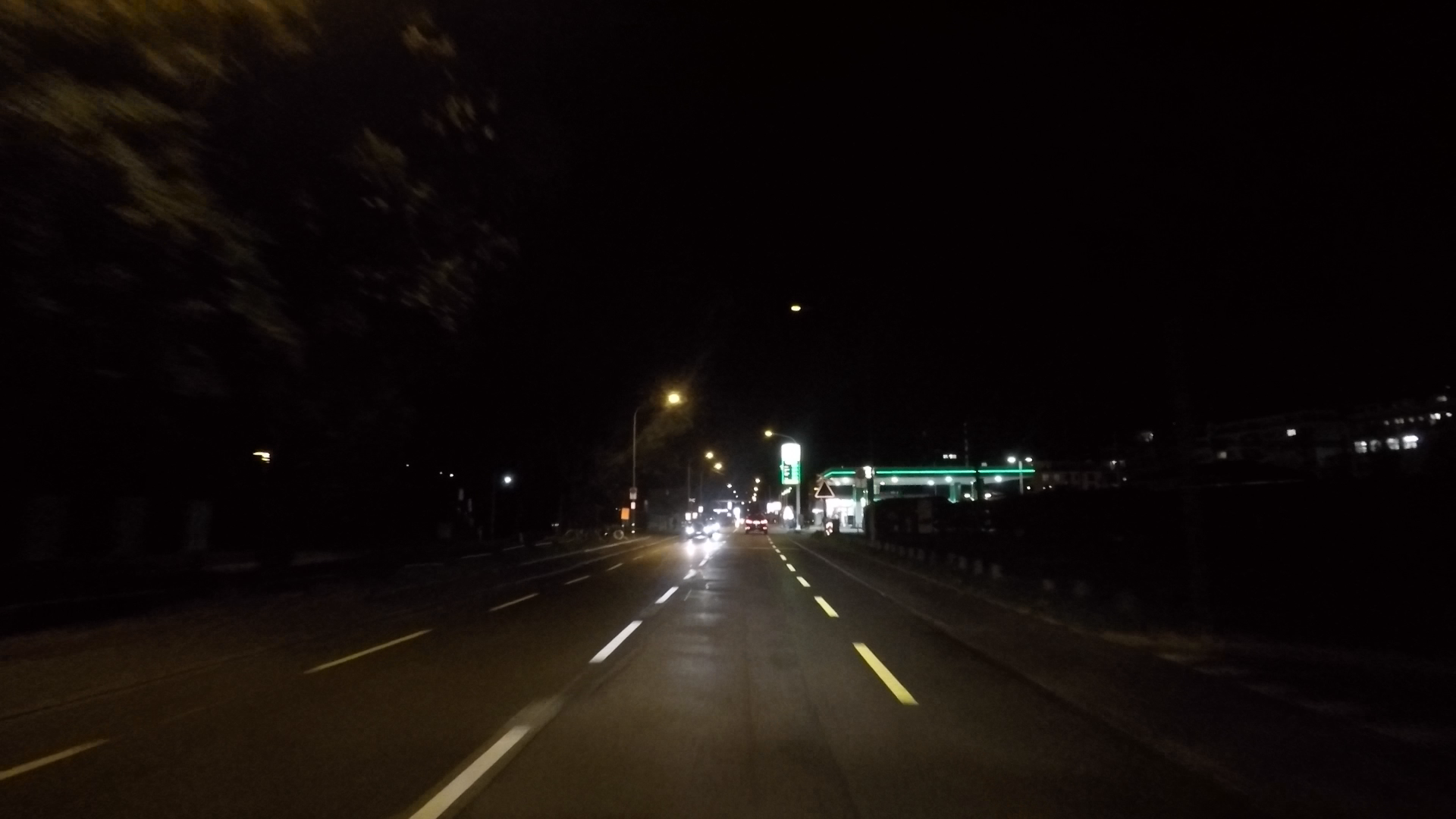}
& \includegraphics[width=0.2\textwidth]{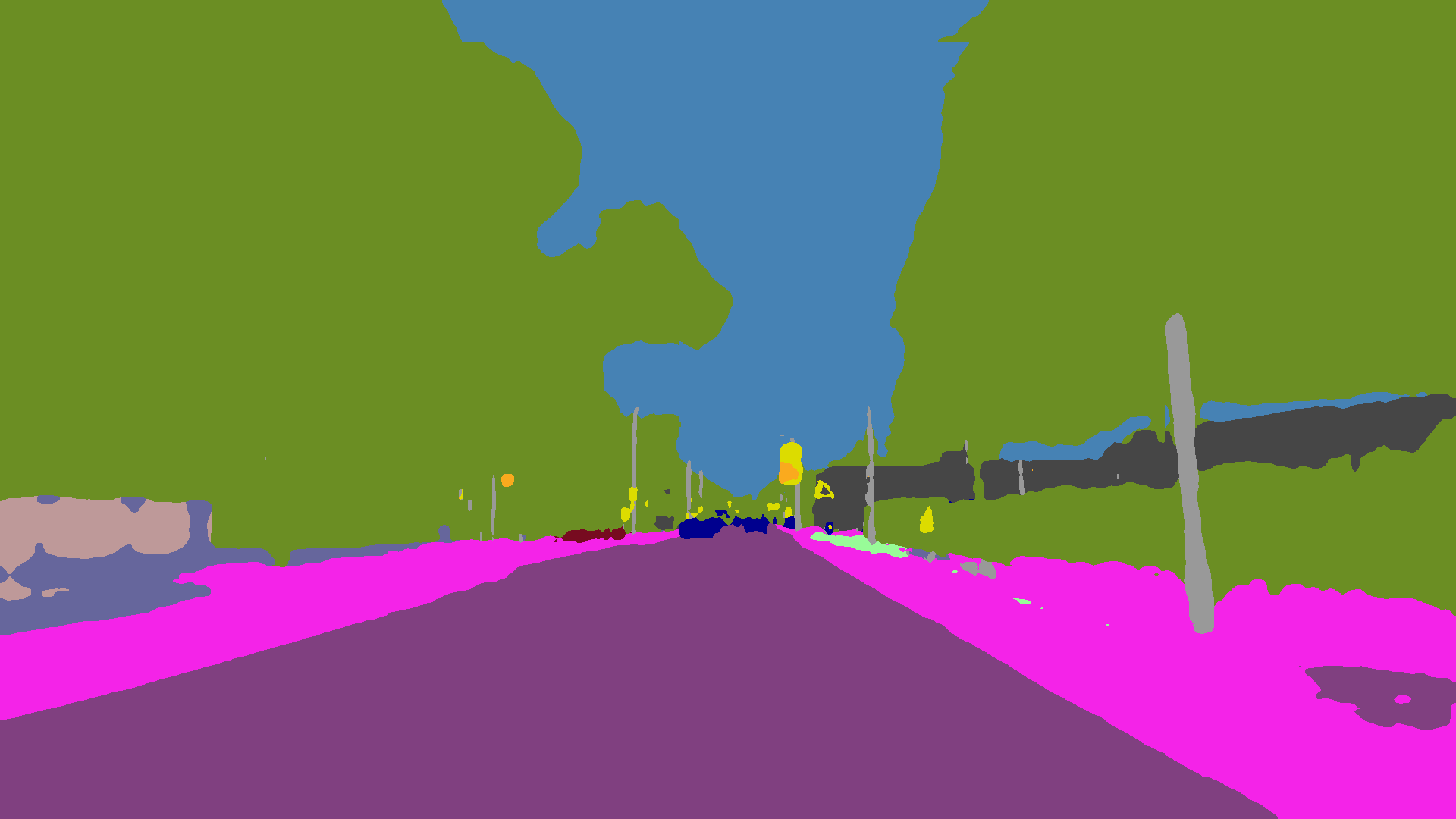}
& \includegraphics[width=0.2\textwidth]{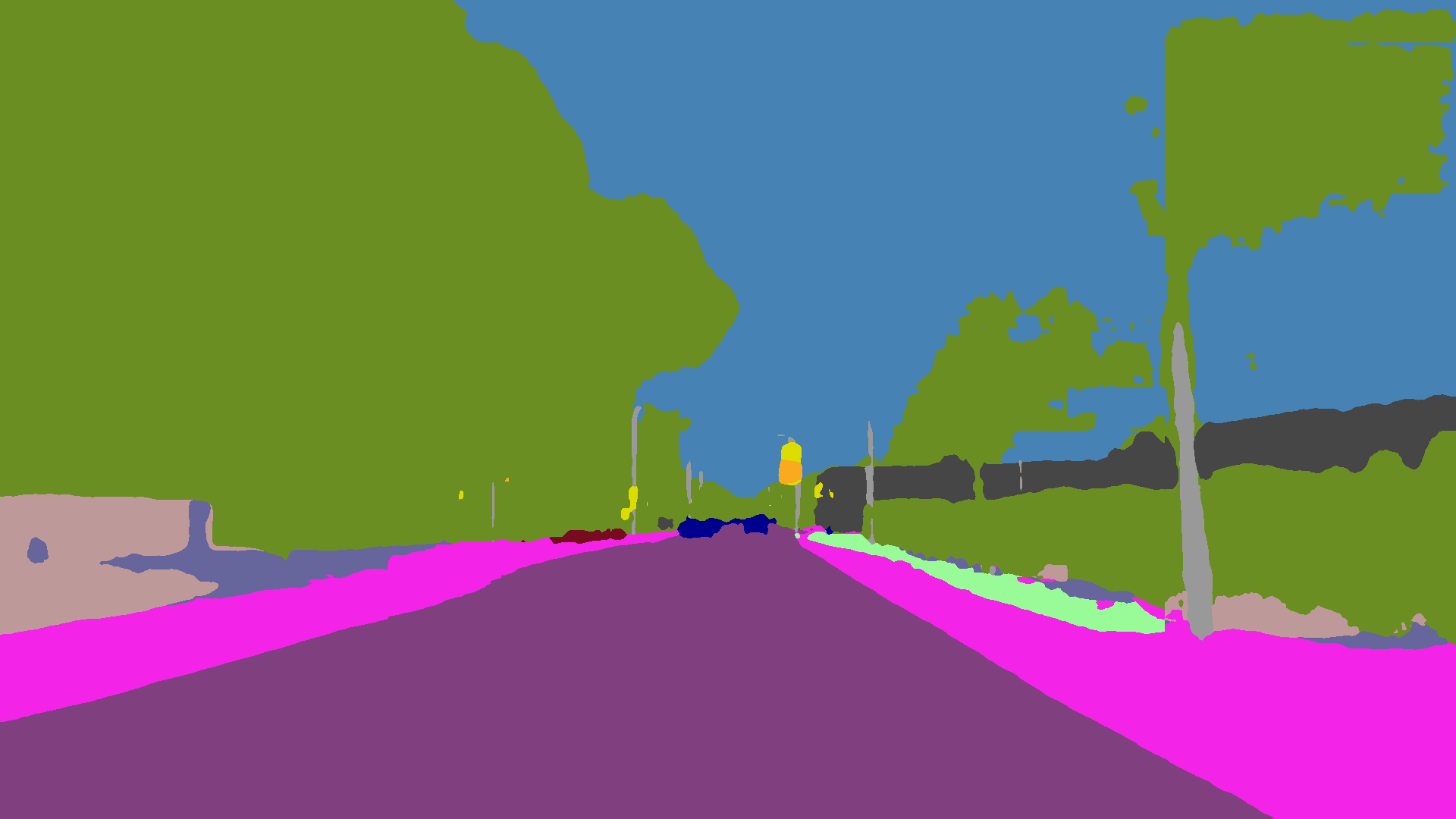}
& \includegraphics[width=0.2\textwidth]{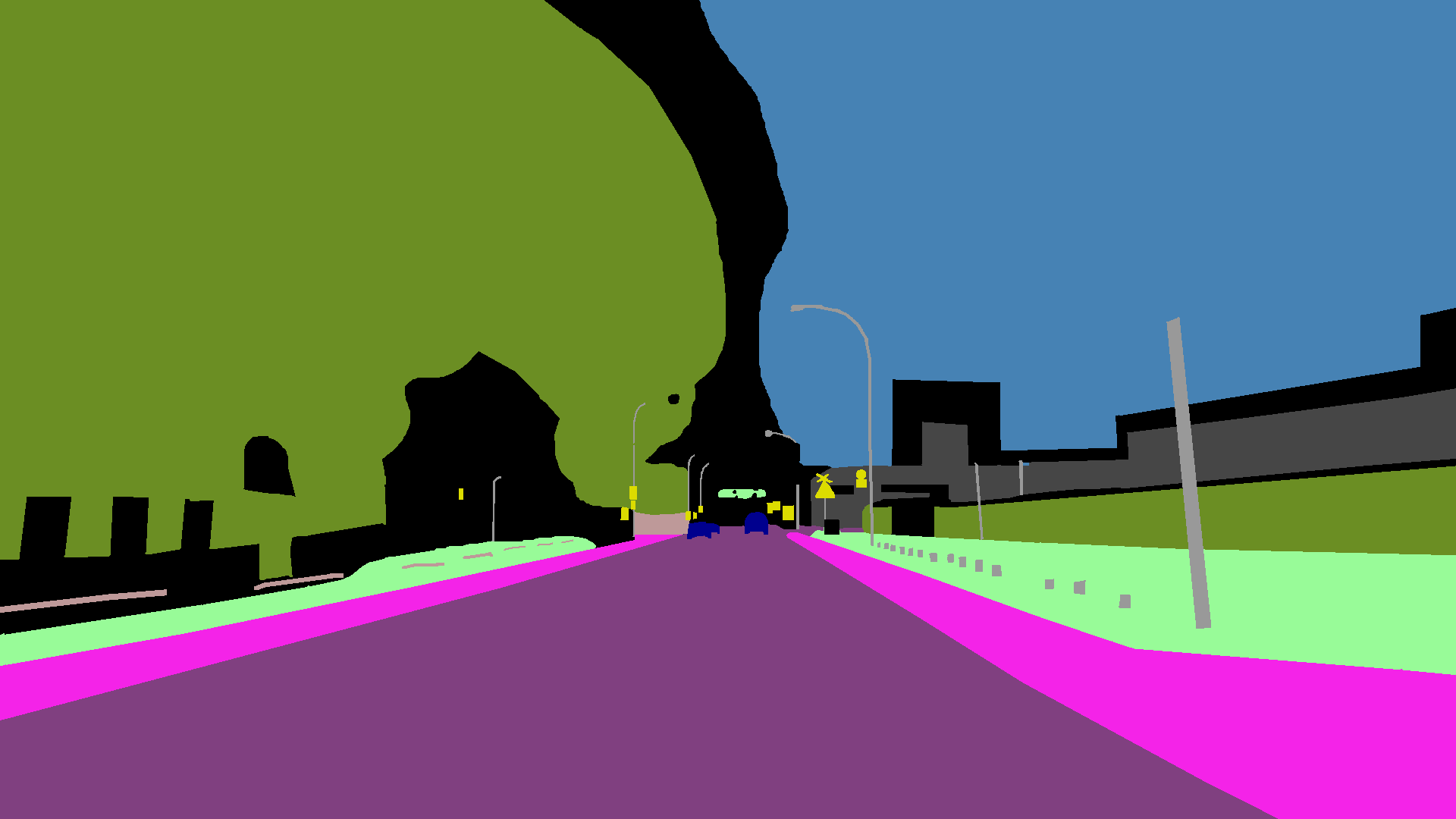} \\

\includegraphics[width=0.2\textwidth]{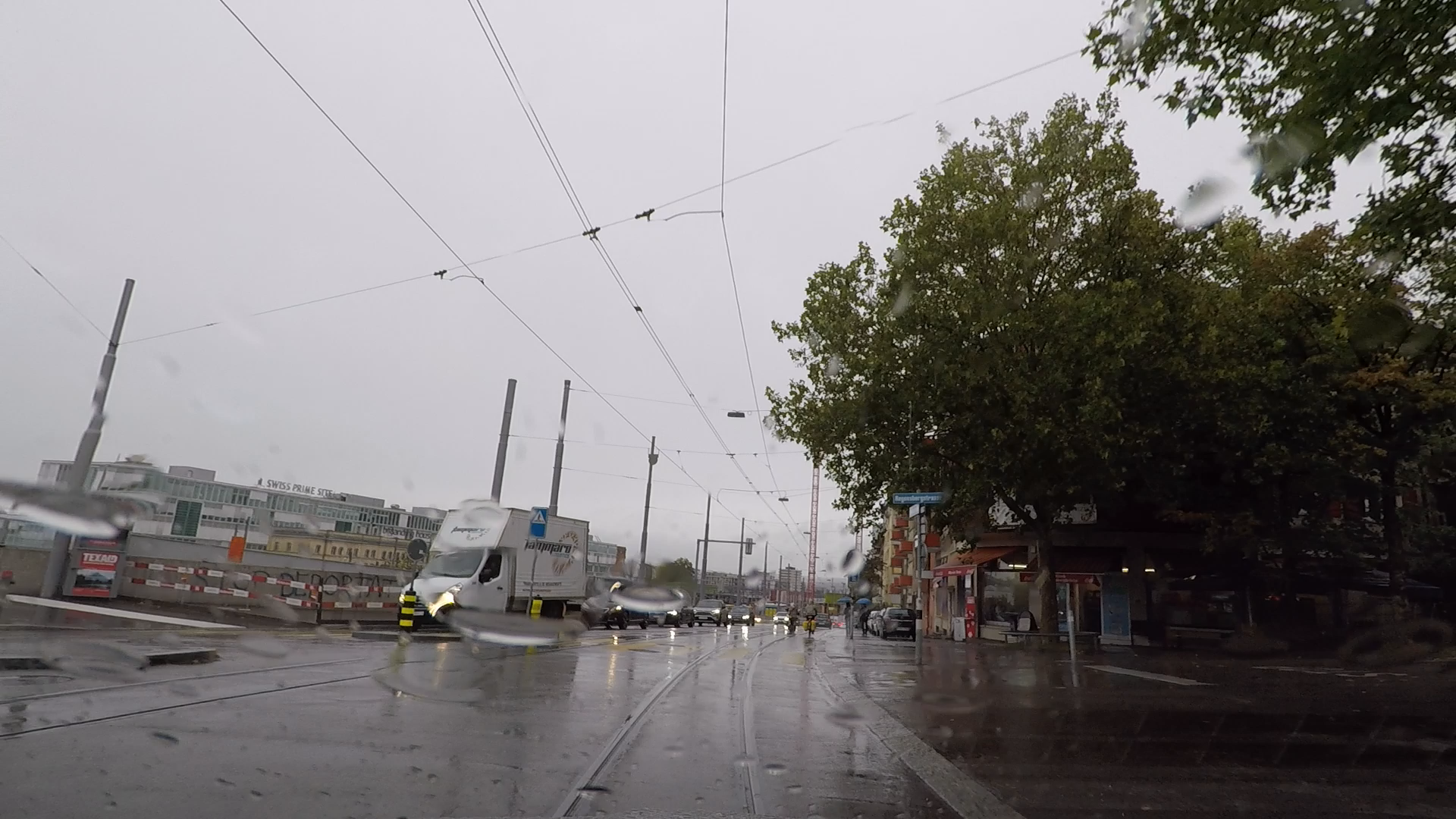}
& \includegraphics[width=0.2\textwidth]{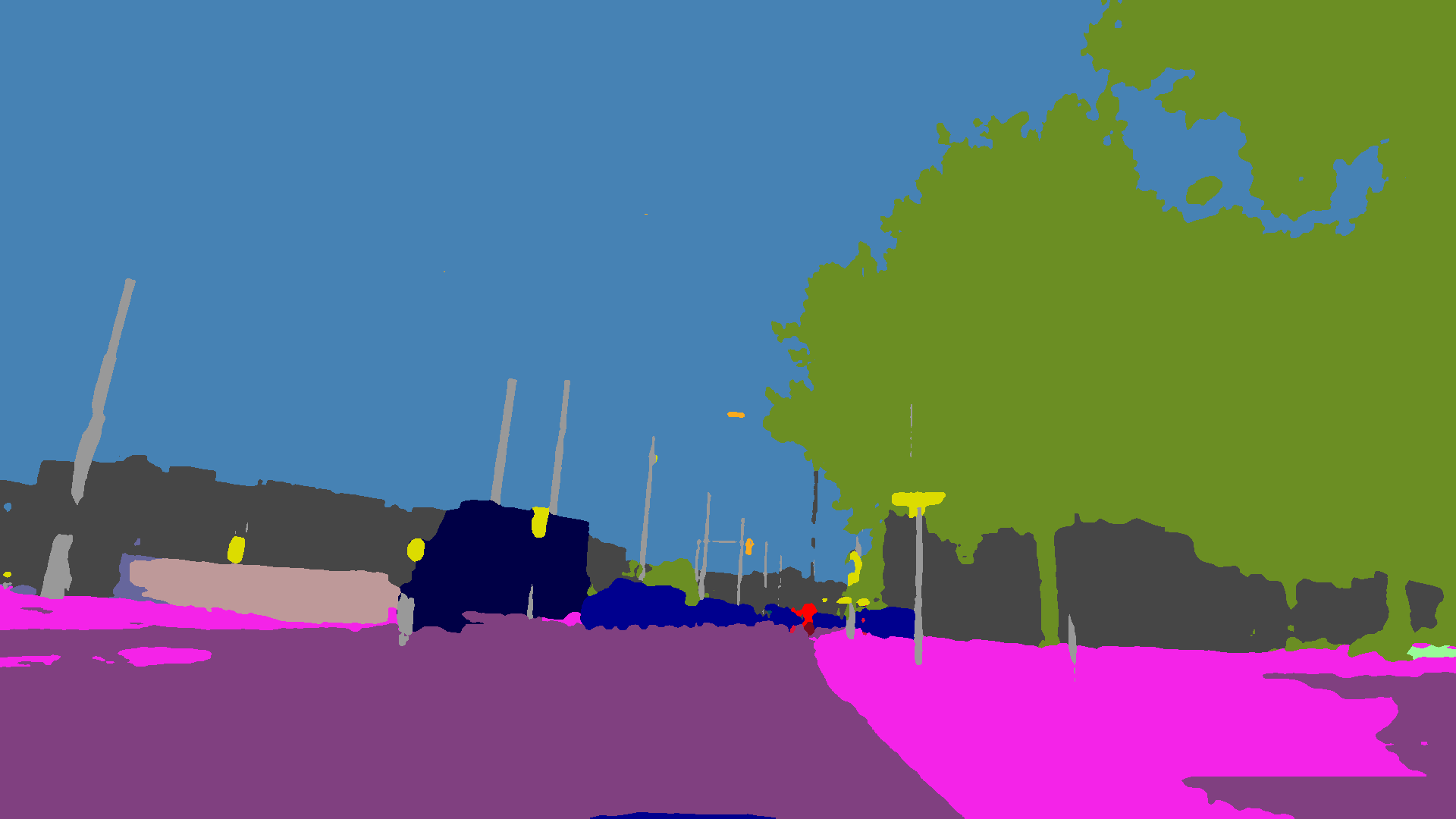}
& \includegraphics[width=0.2\textwidth]{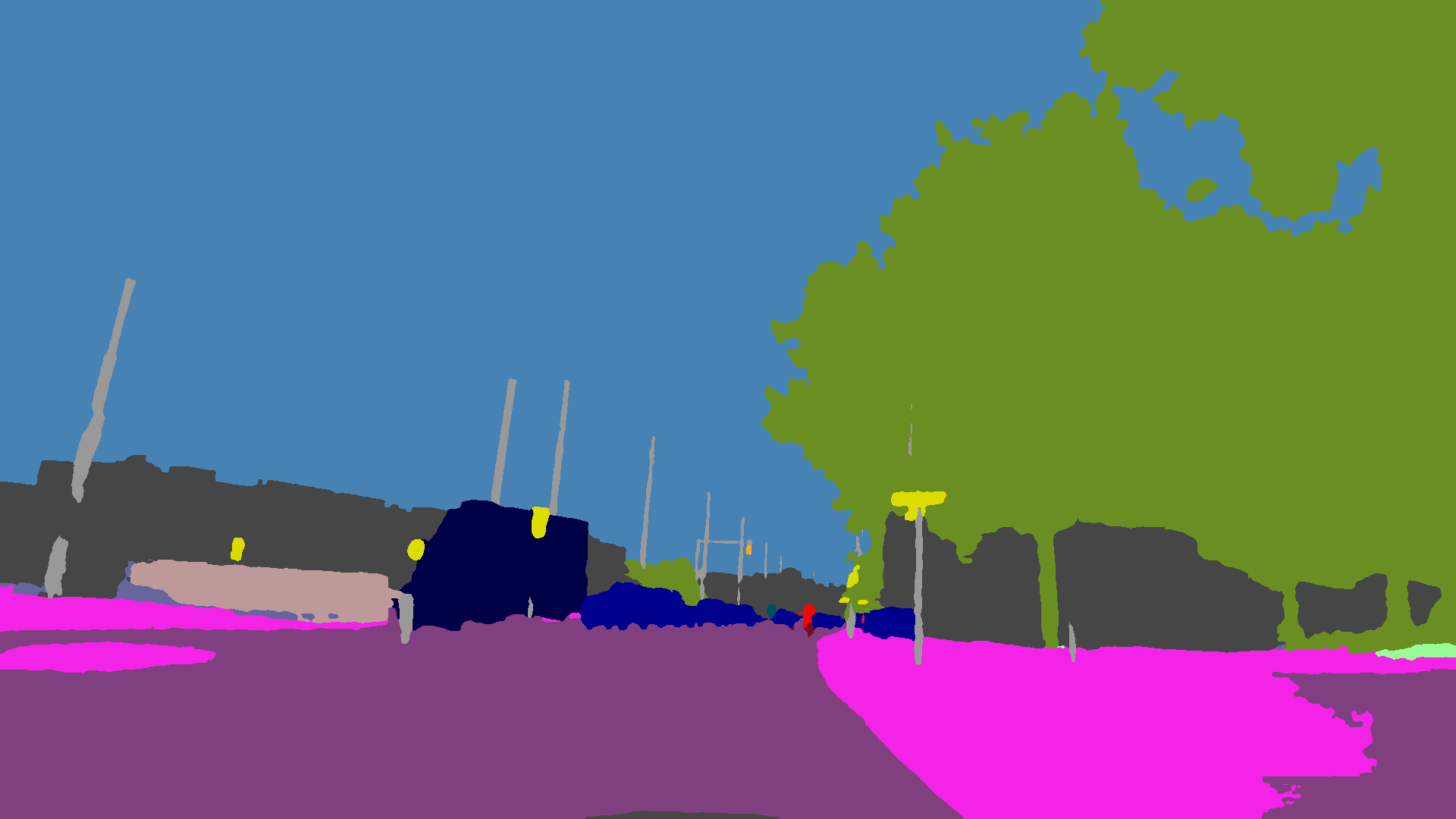}
& \includegraphics[width=0.2\textwidth]{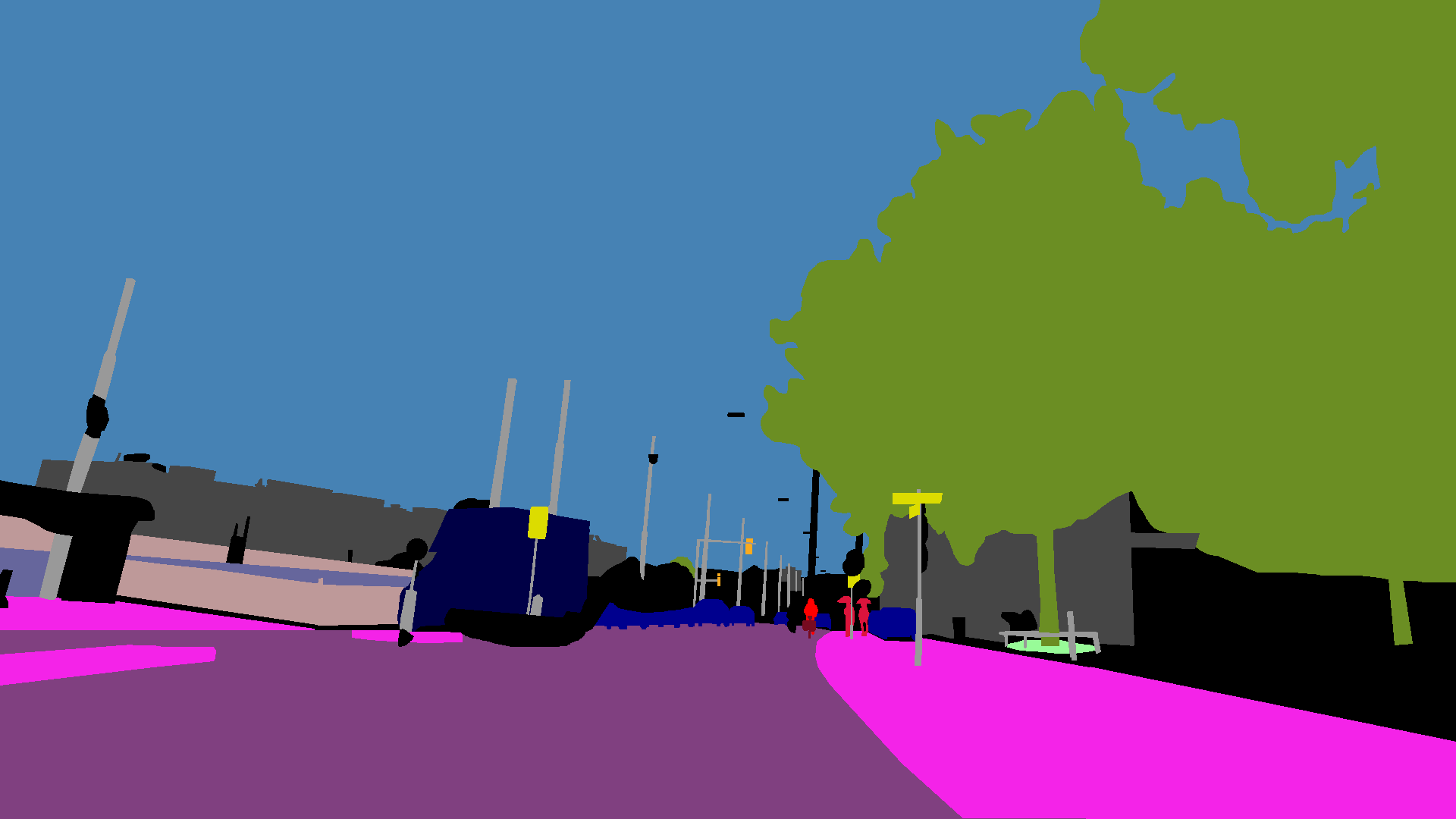} \\

\multicolumn{4}{c}{
    \includegraphics[width=0.8\textwidth]{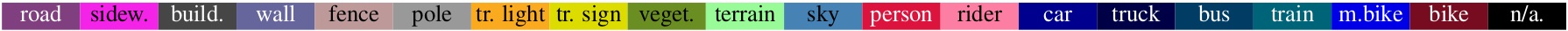}
} \\

\end{tabular}

\caption{Qualitative comparison of ICM with MIC-DINO baseline on CS→ACDC (row 1: snow, row 2: fog, row 3: night, row 4: rain). ICM improves the road segmentation significantly. The class conflation of foggy sky/building, snowy sky/road is mitigated. ICM also segments better with difficult classes, such as sidewalk and traffic light. Better viewed in color and zoomed in.}
\label{fig:qualitative_grid}
\end{figure*}
In this section, we compare the performance  with and without the ICM module using the Cityscapes $\rightarrow$ ACDC UDA benchmark, with two different base models~(MIC-DINO and MIC--Reproduced).

In MIC-DINO, we replace the MIC backbone with DINOv3, due to its strong representation learning capabilities. This improves performance across most semantic classes. However, it exhibits notable class confusion between stuff classes under adverse weather conditions, such as the conflation between sky and road in foggy and snowy weather conditions. Incorporating ICM effectively mitigates this issue, surpassing the baseline MIC-DINO baseline mIoU by 1.3-2.0 pp, with particularly significant gains on road and sky  (see Tab. \ref{tab:baseline}). The qualitative visualization is shown in Fig. \ref{fig:qualitative_grid}. 

We further verify that ICM is not specific to DINO backbone by using the MIC-Reproduced baseline. Due to the unavailability of the official MiT-B5 pretrained checkpoint, we could not exactly reproduce reported MIC results\cite{hoyer2023mic}. Instead, we re-trained MIC under the same setup with MiT-B5 as MIC-Reproduce. The model achieves 65.0 mIoU, while ICM (MiT-B5) achieves 66.5 mIoU. ICM consistently outperforms MIC-Reproduce by +1.5 pp under the MiT-B5 backbone, supporting that gains are method-driven rather than backbone-dependent.

\section{Discussion}
\begin{figure*}[t]
  \centering
  \includegraphics[width=0.8\linewidth]{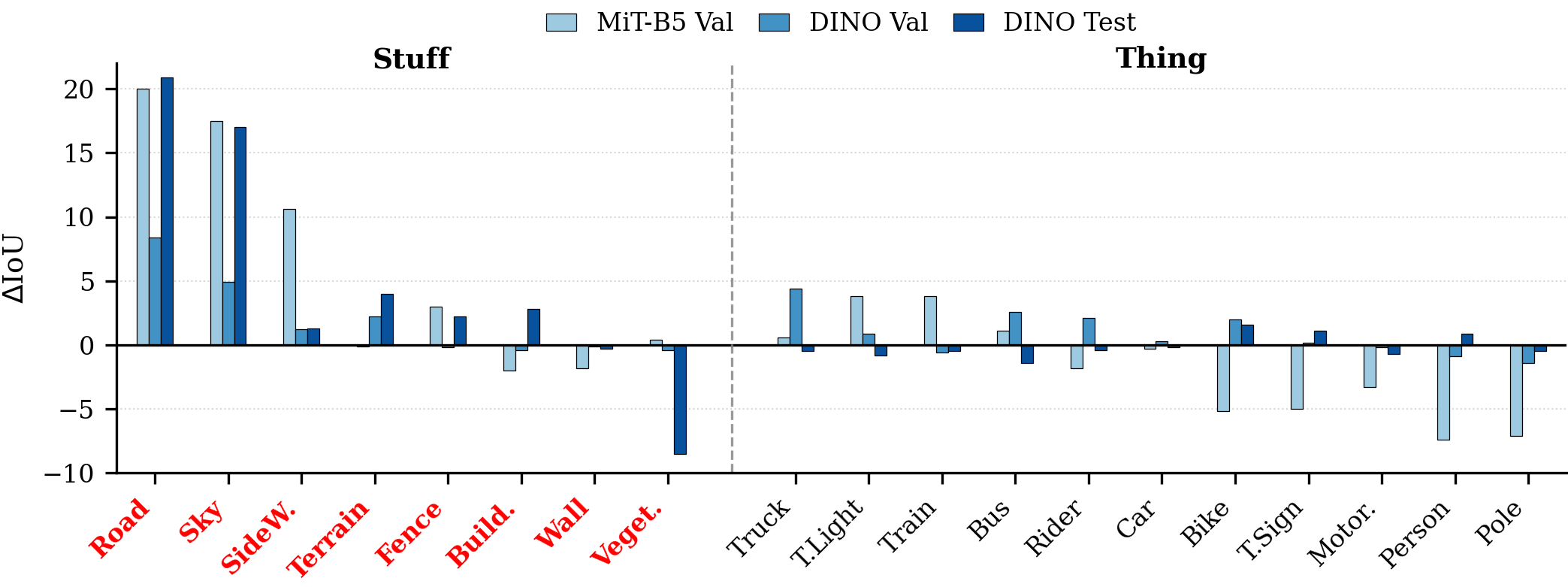}
  \caption{Per-class IoU improvement ($\Delta$IoU = ICM - MIC) on Cityscapes→ACDC for three experimental settings (MiT-B5 Val: Tab.\ref{tab:baseline} group 1, DINO Val: Tab.\ref{tab:baseline} group 2, and DINO Test: Tab.\ref{tab:baseline} group 3). Classes are grouped into stuff and thing. Within each group, classes are sorted by the average improvement across the three settings. }
  \label{fig:per-class-analy}
\end{figure*}

To better understand how ICM affects the performance of individual semantic classes, we analyze the class-wise results in detail. Figure \ref{fig:per-class-analy} illustrates the per-class IoU improvement ($\Delta$IoU = ICM $-$ MIC) on the Cityscapes$\rightarrow$ACDC benchmark under the three experimental settings reported in Table \ref{tab:baseline}. As shown in Fig. \ref{fig:per-class-analy}, the largest performance gains are observed for stuff classes, particularly Road, Sky, and Sidewalk. However, Vegetation exhibits a notable IoU drop in the DINO Test setting, which is discussed in Section \ref{limitation}. Interestingly, despite ICM being applied only to stuff regions, several thing classes also show improved IoU, especially large vehicles such as Truck, Train, and Bus. One possible explanation is that accurate vehicle segmentation benefits from contextual information provided by the surrounding stuff regions. Therefore, improvements in the representation of classes such as roads may indirectly enhance the segmentation performance of nearby vehicles.

\section{Limitation}
The interpretation of the stuff and thing IoU behavior in the loss-region analysis (Tab. \ref{tab:loss_region}) remains preliminary; a representation-level analysis of ICM across different class groups could provide deeper insights into the observed trends(Fig. \ref{fig:per-class-analy}). Although ICM achieves the highest mIoU on Cityscapes $\rightarrow$ ACDC, we suspect that ICM is less effective in severe night conditions, where class confusion (e.g., sky vs.\ vegetation) becomes highly ambiguous. This performance degradation pattern is observed on both the ACDC night subset and the Cityscapes $\rightarrow$ DarkZurich benchmark. In contrast, improvements on ACDC mainly arise from fog and snow conditions, where intra-class consistency is more reliable. Finally, the robustness against different choices of weighting scales ($\lambda_{\mathrm{sup}}$ and $\lambda_{\mathrm{KL}}$) remains to be tested. 
\label{limitation}

\section{Conclusion}
We presented Intra-Class Mixing consistency (ICM), a module for unsupervised domain adaptation in semantic segmentation. It explicitly addresses class conflation under adverse weather. Unlike prior cross-image mixing strategies, ICM performs confusion-guided intra-class patch mixing within the same image, perturbing appearance while preserving semantic identity. With ICM, we establish a new state-of-the-art on Cityscapes $\rightarrow$ ACDC, achieving 75.7\% mIoU and improving upon the previous best result by +1.9 pp. Due to its compatibility with existing UDA frameworks, we hope ICM can contribute to robust adaptation under severe domain shifts. A deeper analysis of how it influences feature representations across stuff and thing classes remains an interesting direction for future work.

\bibliographystyle{splncs04}
\bibliography{main}
\end{document}